%% file: main.tex
\documentclass[10pt]{article} % For LaTeX2e
\usepackage[preprint]{tmlr}

\input{math_commands.tex}

\usepackage{hyperref}
\usepackage{url}

\usepackage[utf8]{inputenc} % allow utf-8 input
\usepackage[T1]{fontenc}    % use 8-bit T1 fonts
\hypersetup{colorlinks=true,citecolor=blue}
\usepackage{booktabs}       % professional-quality tables
\usepackage{amsfonts}       % blackboard math symbols
\usepackage{mathtools}
\usepackage{algorithm}
\usepackage{nicefrac}       % compact symbols for 1/2, etc.
\usepackage{microtype}      % microtypography
\usepackage{xcolor}         % colors
\usepackage{bm}
\newcommand*{\G}{\mathbf{G}}
\newcommand*{\F}{\mathbf{F}}
\newcommand*{\A}{\mathbf{A}}
\newcommand*{\B}{\mathbf{B}}
\newcommand*{\mub}{\bm{\mu}}
\newcommand*{\gamb}{\bm{\gamma}}
\newcommand*{\thetab}{\bm{\theta}}
\newcommand*{\Psib}{\bm{\Psi}}
\newcommand*{\Phib}{\bm{\Phi}}
\newcommand*{\Sigmab}{\bm{\Sigma}}
\newcommand*{\Omegab}{\bm{\Omega}}

\title{Supervised Quadratic Feature Analysis: Information Geometry for Dimensionality Reduction}

\author{\name Daniel Herrera-Esposito \email dherresp@sas.upenn.edu \\
      \addr Department of Psychology\\ 
      University of Pennsylvania
      \AND
      \name Johannes Burge \email jburge@psych.upenn.edu \\
      \addr Department of Psychology\\ 
      University of Pennsylvania
      }

\def\month{06}  % Insert correct month for camera-ready version
\def\year{2026} % Insert correct year for camera-ready version
\def\openreview{\url{https://openreview.net/forum?id=jwNJiLphnZ}} % Insert correct link to OpenReview for camera-ready version

\begin{document}

\maketitle

\begin{abstract}
Supervised dimensionality reduction maps labeled data into a low-dimensional feature space while preserving class separation. A common strategy is to learn features that maximize a measure of statistical dissimilarity between the class-conditional probability distributions. Information geometry, which is rooted in Riemannian geometry, provides an alternative framework for measuring class dissimilarity. It treats probability distributions as points in a statistical manifold and uses the Fisher information metric to define a geodesic distance--the Fisher-Rao distance--between distributions The Fisher-Rao distance is an appealing candidate for measuring class separation because the Fisher information metric is a local measure of discriminability, and because it allows a geometric interpretation. Here, we present Supervised Quadratic Feature Analysis (SQFA), a supervised dimensionality reduction method which learns linear features that maximize Fisher-Rao distances between class-conditional distributions, under Gaussian assumptions. In multiple real world datasets, we find that SQFA features support classification accuracy that is competitive with features that maximize more popular measures of dissimilarity, or that are learned by other state-of-the-art dimensionality reduction methods. Notably, the best classification accuracy is achieved by SQFA-H features, a variant of SQFA that maximizes the Hellinger distance, a rarely used objective for dimensionality reduction. These results demonstrate the potential of information geometry as a tool for supervised dimensionality reduction. We provide a Python implementation of SQFA at \url{https://github.com/dherrera1911/sqfa}.
\end{abstract}

\section{Introduction}
\label{sec:intro}

\begin{figure*}[t]
\begin{center}
\centerline{\includegraphics[width=0.9\textwidth]{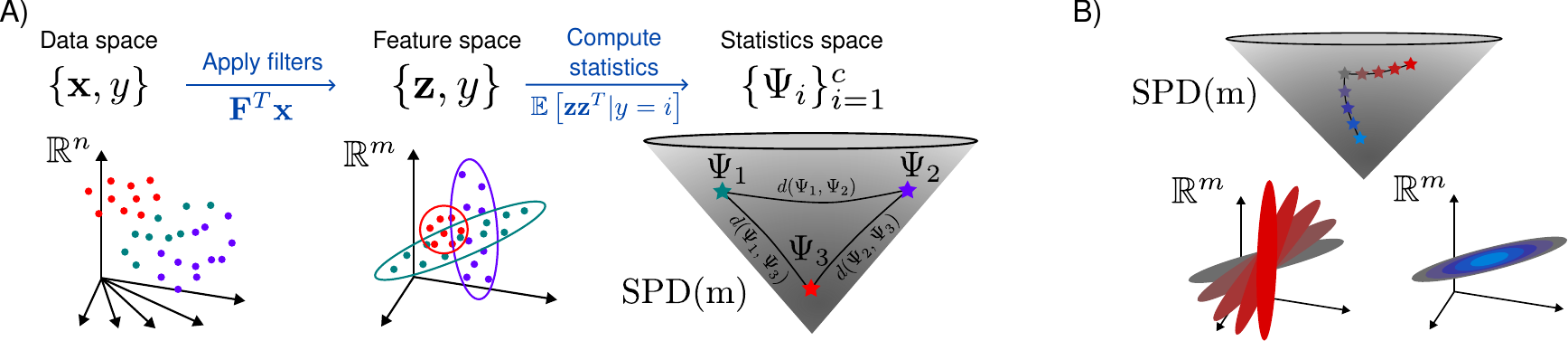}}
\caption{SQFA learns features using information geometry.
  \textbf{A)} SQFA and smSQFA map the $n$-dimensional data into an $m$-dimensional
  feature space using the linear filters $\F$. In smSQFA, the
  class-specific second-moment matrices of the features
  are represented as points in the $\mathrm{SPD(m)}$ manifold.
  Learning is achieved by maximizing the Fisher-Rao distances between
  classes in $\mathrm{SPD(m)}$.
  \textbf{B)} Each point in $\mathrm{SPD(m)}$ (top) corresponds to
  a second-moment ellipse (below). As the distance in $\mathrm{SPD(m)}$ increases,
  the second-order statistics become more different and the
  distributions more discriminable.}
\label{fig:geometry}
\end{center}
\end{figure*}

Consider a random vector $\mathbf{x}\in \mathbb{R}^n$
with label $y \in \{1, \ldots, c\}$, where $c$ is the number
of classes. Supervised dimensionality reduction aims to
map the high-dimensional variable $\mathbf{x}$
to a lower-dimensional variable
$\mathbf{z} \in \mathbb{R}^m$ that best supports
classification performance, or class separation.
There are many methods for learning nonlinear
features, like deep neural networks \citep{lecun_deep_2015}
and manifold learning \citep{sainburg_parametric_2021}.
However, methods for learning linear features of the form
$\mathbf{z} = \F^T\mathbf{x}$ are still in 
demand because of their simplicity, interpretability,
and data efficiency \citep{cunningham_linear_2015}. 
A common approach to linear supervised dimensionality reduction is
to learn features that maximize a statistical or information-theoretic
measure of dissimilarity between pairwise class distributions \citep{fukunaga_introduction_1990},
like the squared Mahalanobis distance (i.e. Linear Discriminant Analysis; LDA),
the Bhattacharyya distance \citep{choi_feature_2003},
or the KL divergence \citep{tao_general_2007,dwivedi_discriminant_2022}.

Information geometry provides an alternative approach for measuring the
dissimilarity between probability distributions, by considering them
as points in a statistical manifold. It uses the Fisher information metric,
a principled measure of local
%discriminability\footnote{When the Fisher information of
%$p(\mathbf{z}|\thetab)$ in the direction $\thetab'$ is large,
%the variance of the unbiased estimate of $\thetab$ along that direction has a smaller lower
%bound (Cramér-Rao bound) i.e. $\thetab$ can be estimated more precisely. Then, two distributions
%with parameters $\thetab$ and $\thetab + \epsilon \thetab'$
%can be better discriminated from the samples of $p(\mathbf{z}|\thetab)$. This is
%what is meant here by local discriminability.},
discriminability\footnote{We use the term discriminability loosely to refer to
the ability to discriminate between two distributions based on observed samples.
Local discriminability of $p(\mathbf{z}|\thetab)$ in the direction $\thetab'$ refers to
the ability to discriminate between $p(\mathbf{z}|\thetab)$ and
$p(\mathbf{z}|\thetab+ \epsilon \thetab')$, where $\epsilon \thetab'$ is a small perturbation.
Local discriminability is closely related to Fisher information via the Cramér-Rao bound}
commonly used in machine learning and
neuroscience \citep{dayan_theoretical_2005,amari_information_2016},
to measure distances in the manifold of probability distributions.
The geodesic distance induced by this metric is the Fisher-Rao distance.

Maximizing Fisher-Rao distances between classes is therefore
an attractive learning objective for several reasons.
First, there has been a recent surge of interest in the geometry of
neural representations in both machine learning and neuroscience \citep{kriegeskorte_neural_2021,chung_neural_2021}.
The Fisher information metric is commonly used for characterizing
local representation geometry, but it is limited to infinitesimally close distributions
\citep{wang_geometry_2021,ding_information_2023,
feather_discriminating_2024,ye_speed_2024,zhou_unified_2024}.
The Fisher-Rao distance provides a tool for studying the non-local geometry
of neural representations \citep{kriegeskorte_neural_2021}.
Second, a long-standing hypothesis from perception science dating back to
Riemann himself proposes that geodesic distances induced by perceptual discriminability
metrics (analogous to the Fisher information metric) can capture
supra-threshold (i.e. non-local) perceptual judgments of similarity
\citep{riemann_ueber_1867,fechner_elemente_1860,vacher_perceptual_2024,zhou_unified_2024,hong_comprehensive_2025}.
This makes the Fisher-Rao distance an appealing candidate for learning
representations that maximize supra-threshold perceptual dissimilarities.
Third, given the broad interest in information geometry,
it is of standalone interest to understand how maximizing Fisher-Rao
distances compares to other common objective functions in machine learning.

Despite its theoretical and practical appeal, very little work
has studied the use of Fisher-Rao distances as an objective for learning
\citep{miyamoto_closed-form_2024}. To our knowledge,
Fisher-Rao distances between classes have not been used for dimensionality reduction.
One possible reason is that closed-form expressions for the
Fisher-Rao distance are not available for general multivariate Gaussians
\citep{nielsen_simple_2023}.
Another possible reason is that information-theoretic measures of dissimilarity
(e.g.\ Bhattacharyya distance, KL divergence)
are more directly linked to classification error, making them
a more obvious choice \citep{fukunaga_introduction_1990}.

Here, we present Supervised Quadratic Feature Analysis (SQFA),
a dimensionality reduction method that learns linear features by maximizing Fisher-Rao
distances between class-conditional distributions, under Gaussian assumptions.
We use an exact expression for the Fisher-Rao distance for zero-mean
Gaussians, and the Calvo-Oller lower bound \citep{calvo_distance_1990}
as a closed-form approximation in the general case. We compare the
classification performance supported by SQFA features to that supported by features
that maximize other measures of dissimilarity.
We also compare performance to other popular methods for dimensionality reduction.
Our analyses show that:
\begin{itemize}
\item Maximizing Fisher-Rao distances between classes leads to features that
  support classification accuracy competitive with state-of-the-art
  methods for QDA and kNN classifiers.
\item Maximizing the Calvo-Oller bound is a practical way to maximize the Fisher-Rao distance.
\item Features learned by maximizing the Hellinger distance (SQFA-H) consistently support
  the highest classification accuracy. SQFA-H is a state-of-the-art method for learning linear
    features that maximize classification accuracy.
\end{itemize}

\section{Fisher-Rao distance as a discriminability proxy}

\paragraph{Notation.}
The class-conditional means and covariances of the data
variable $\mathbf{x}$ are denoted by
\mbox{$\gamb_i = \mathbb{E}\left[\mathbf{x}|y=i\right]$} and
\mbox{$\Phib_i = \mathbb{E}\left[(\mathbf{x}-\gamb_i)(\mathbf{x}-\gamb_i)^T|y=i\right]$},
respectively.
The low-dimensional projection of $\mathbf{x}$
is given by $\mathbf{z} = \F^T\mathbf{x}$, where
$\F \in \mathbb{R}^{n \times m}$ is a matrix of
filters, and the respective class-conditional means,
second moments, and covariances are given by
$\mub_i = \mathbb{E}\left[\mathbf{z}|y=i\right]$,
$\Psib_i = \mathbb{E}\left[\mathbf{z}\mathbf{z}^T|y=i\right]$,
and $\Sigmab_i = \Psib_i - \mub_i\mub_i^T$.
We also denote by $\thetab_i$ the parameters of
$p(\mathbf{z}|y=i)=p(\mathbf{z}|\thetab_i)$, which in the
Gaussian cases is $\thetab_i = (\mub_i, \Sigmab_i)$.

\subsection{Supervised dimensionality reduction via dissimilarity maximization}
\label{sec:geometry_setup}

The goal of linear supervised dimensionality reduction 
is to find filters $\F$ such that the classes
are as discriminable as possible in the low-dimensional
space of the variable $\mathbf{z} = \F^T\mathbf{x}$.
A common way to achieve this is to maximize a dissimilarity
measure between the class-conditional distributions.
For Gaussian-distributed classes
$p(\mathbf{z}|y=i) = \mathcal{N}(\mub_i, \Sigmab_i)$,
given a dissimilarity measure $d(\thetab_i, \thetab_j)$ where
$\thetab_i = (\mub_i, \Sigmab_i)$, the problem reduces to
\begin{equation}
\label{eq:loss1}
\underset{{\F\in\mathbb{R}^{n \times m}}}\argmax \hspace{0.2em}
  \sum_{i=1}^{c}\sum_{j=1}^{c} d(\thetab_i,\thetab_j)
\end{equation}
If $d(\thetab_i, \thetab_j)$ is a good proxy for discriminability,
the learned features should support high classification accuracy
(Figure~\ref{fig:geometry}). We argue that
the Fisher-Rao distance, $d_{FR}(\thetab_i, \thetab_j)$,
is a sensible proxy for discriminability.

\subsection{Fisher-Rao distance as accumulated local discriminability}
\label{sec:fr_distance}

For a distribution $p(\mathbf{z}|\thetab)$, the Fisher information
in the direction $\thetab'$ in the parameter space is defined as
\begin{equation}
  \mathcal{I}_{\thetab}(\thetab') = \mathbb{E}\left[(s(\mathbf{z},\thetab) \cdot \thetab')^2\right] =
  \thetab'^T \mathbf{J}(\thetab) \thetab'
\end{equation}
where the expectation is over $\mathbf{z}$,
$s(\mathbf{z},\thetab) = \nabla_{\thetab} \log p(\mathbf{z}|\thetab)$
is the score function, and $\mathbf{J}(\thetab) = \mathbb{E}[s(\mathbf{z},\thetab)s(\mathbf{z},\thetab)^T]$ is the Fisher information matrix.
The quantity $\sqrt{\mathcal{I}_{\thetab}(\thetab')}$ measures the
discriminability between $p(\mathbf{z}|\thetab)$ and
$p(\mathbf{z}|\thetab + \epsilon \thetab')$,
where $\epsilon\thetab'$ is a small perturbation.

Let $\thetab(t)=(\mub(t),\Sigmab(t))$ be the Fisher-Rao geodesic (i.e.\ the shortest curve)
connecting $\mathcal{N}(\mub_i,\Sigmab_i)$ to $\mathcal{N}(\mub_j,\Sigmab_j)$
along the manifold of Gaussian distributions,
where $\thetab(0) = \thetab_i$, $\thetab(1) = \thetab_j$,
and $\thetab'(t)$ is the velocity. The Fisher-Rao distance is
obtained by integrating the speed along the geodesic,
\mbox{$d_{FR}(\thetab_i, \thetab_j) = \int_0^1 \|\thetab'(t)\| dt$}.
The Fisher information metric defines the speed
as $\|\thetab'\| = \sqrt{\mathcal{I}_{\thetab}(\thetab')}$, which
for the Gaussian case is
\begin{align}
  \|\thetab'(t)\|
    &= \sqrt{\mathcal{I}_{\thetab(t)}(\thetab'(t))} 
    = \left[ \mub'(t)^T \Sigmab(t)^{-1} \mub'(t) 
    + \tfrac{1}{2}\,\text{Tr}\left(\Sigmab(t)^{-1}\Sigmab'(t)
                         \Sigmab(t)^{-1}\Sigmab'(t)\right) \right]^{1/2}
\end{align}
Then, $\|\thetab'(t)\|$ is a measure of the discriminability between
$\mathcal{N}(\mub(t),\Sigmab(t))$ and
$\mathcal{N}(\mub(t+\mathrm{d}t),\Sigmab(t+\mathrm{d}t))$. Thus, the
Fisher-Rao distance can be expressed as
$d_{FR}(\thetab_i, \thetab_j) =
\int_0^1 \sqrt{\mathcal{I}_{\thetab(t)}\left(\thetab'(t)\right)} dt$,
and it can be conceptualized as
the accumulated discriminability of the infinitesimal perturbations
transforming $\mathcal{N}(\mub_i,\Sigmab_i)$ into
$\mathcal{N}(\mub_j,\Sigmab_j)$
along the geodesic, making it a sensible proxy for discriminability.

\subsection{Closed-form expressions for the Fisher-Rao distance}

There is no closed-form expression for the Fisher-Rao distance between
arbitrary Gaussians.  Numerical methods exist to compute it
\citep{nielsen_simple_2023,nielsen_pybregman_2024}, but they are
too costly to be used as optimization objectives. To overcome this problem,
we use two complementary approaches: 1) we consider
the special case of zero-mean Gaussians, and
2) we use a closed-form lower bound for the general case.

\paragraph{Zero-mean Gaussians.} The manifold of $m$-dimensional zero-mean
Gaussians is equivalent to the manifold of $m$-by-$m$ symmetric positive
definite matrices, $\mathrm{SPD(m)}$, where each point corresponds to a
covariance matrix. The Fisher-Rao distance between zero-mean Gaussians,
denoted $d_{FR}(\Sigmab_i,\Sigmab_j)$, is proportional to the
affine-invariant distance $d_{AI}$ in $\mathrm{SPD(m)}$
\citep{atkinson_raos_1981}
\begin{align}
\label{eq:ai_dist}
  d_{AI}(\Sigmab_i,\Sigmab_j) 
  & =  \sqrt{\sum_{k=1}^{m} \log^2 \lambda_k} =
 d_{FR}(\Sigmab_i,\Sigmab_j) \sqrt{2} 
\end{align}
where $\lambda_k$ is the $k$-th generalized eigenvalue of the pair $(\Sigmab_i,\Sigmab_j)$.
Leveraging this closed-form expression, we define smSQFA (for `second-moment SQFA'),
which assumes zero-mean Gaussians, and that maximizes pairwise
$d_{AI}(\Psib_i,\Psib_j)$, where $\Psib_i$ is the second-moment matrix for class $i$ in the feature space.

Zero-mean distributions are relevant for some applications, like EEG
recordings \citep{horev_geometry-aware_2017}, some local visual
tasks \citep{burge_image-computable_2020}, and auditory waveform data, among others.
Also, the zero-mean case is more tractable and allows one to draw direct links
between the Fisher-Rao distance and measures of discriminability, such as
the ratio between class-conditional feature variances (see Appendix~\ref{apd:zero_mean}).
However, our main motivation for considering the zero-mean case is that it allows
us to maximize the exact Fisher-Rao distance, which is important for understanding
the link between maximizing Fisher-Rao distances and discriminability
without the confound of using an approximation (see below).

\paragraph{Calvo-Oller bound.} For arbitrary Gaussians, we use the Calvo-Oller
bound \citep{calvo_distance_1990}. It is obtained by embedding
$\thetab$ into $\mathrm{SPD(m+1)}$ as
\begin{equation}
  \label{eq:embedding}
    \Omegab = \begin{bmatrix}
      \Sigmab + \mub \mub^T & \mub \\
      \mub^T & 1 \\
    \end{bmatrix}
\end{equation}
The Calvo-Oller formula is given by
$d_{CO}(\thetab_i, \thetab_j) = d_{AI}(\Omegab_i,\Omegab_j)/\sqrt{2}$,
and it provides a lower bound for $d_{FR}(\thetab_i, \thetab_j)$.
This bound has desirable properties: it is a true distance,
it matches $d_{FR}$ when $\mub_i = \mub_j$ (Appendix~\ref{apd:calvo_oller}),
and it is invariant to invertible linear transformations
of $\mathbf{z}$ \citep{nielsen_simple_2023}.
SQFA learns filters by maximizing the Calvo-Oller bound between classes
(Algorithm~\ref{alg:sqfa}).
Empirically, we found that the Calvo-Oller bound is a good proxy
for the Fisher-Rao distance in the analyzed datasets. 
It is also worth noting that the Calvo-Oller bound
can be easily extended to other elliptical distributions
(e.g.\ multivariate t-Student and Cauchy
distributions) \citep{calvo_distance_2002,nielsen_simple_2023},
which may facilitate future work that extends SQFA to non-Gaussian distributions.

\section{Related work}
\label{sec:related}

\subsection{Parametric methods maximizing dissimilarity between class-conditional distributions}
\label{sec:related_work_parametric}

Maximizing a measure of dissimilarity between class-conditional distributions
is an established approach for supervised dimensionality reduction.
For example, the canonical method LDA can be interpreted as 
maximizing pairwise squared Mahalanobis distances (Appendix~\ref{apd:lda}).
However, the Mahalanobis distance is limited by the assumption that all
class-conditional distributions have identical covariance.
To overcome this limitation, other methods propose using other measures
of dissimilarity. Under Gaussian assumptions the most popular are
the Chernoff distance and its special case,
the Bhattacharyya distance\footnote{Neither the Bhattacharyya nor the Chernoff
distances are a true distance, since they do
not satisfy the triangle inequality. Nonetheless, we refer to them as
distances as is common in the literature.}
\citep{duin_linear_2004,rueda_linear_2008,choi_feature_2003}. 
One reason for the success of these distances is that they provide an upper bound
for the Bayes classification error 
\citep{kailath_divergence_1967,fukunaga_introduction_1990}.
Another reason is that they have simple closed-form expressions.

The Bhattacharyya distance between two distributions
$p(\mathbf{z})$ and $q(\mathbf{z})$ is given by
$d_{B}(p,q) = -\log BC(p,q)$, where
$BC(p,q) = \int \sqrt{p(\mathbf{z})q(\mathbf{z})} d\mathbf{z}$
is the dot product between the square-root densities.
For two Gaussians with parameters $\thetab_i$ and $\thetab_j$,
denoting $\Sigmab = (\Sigmab_i + \Sigmab_j)/2$,
the distance is given by
\begin{align}
  \label{eq:bhattacharyya}
  d_{B}(\thetab_i,\thetab_j) & = \frac{1}{8} (\mub_i-\mub_j)^T \Sigmab^{-1}  (\mub_i-\mub_j) 
  + \frac{1}{2} \log \frac{\det\left(\Sigmab\right)}{\sqrt{\det\left(\Sigmab_i\right)\det\left(\Sigmab_j\right)}}
\end{align}
Another important measure of dissimilarity is the Hellinger distance,
defined as the Euclidean distance between square-root densities,
$d_{H}(p,q) =  \sqrt{\frac{1}{2} \int (\sqrt{p(\mathbf{z})} - \sqrt{q(\mathbf{z})})^2} d\mathbf{z}$.
It relates to the Bhattacharyya distance via $d_{H}(p,q) = \sqrt{1 - e^{-d_B(p,q)}}$.
Therefore, it also has a closed-form expression for Gaussians,
and provides the same bound on the Bayes classification error.
It is a true distance, bounded between 0 and 1, and it converges to
the Fisher-Rao distance for infinitesimally close distributions \citep{amari_information_2016}. 
Notably, $d_H$ has not been used for dimensionality reduction in the multiclass
Gaussian case (see \cite{carter_information_2009} for a non-parametric
example, and \cite{dwivedi_discriminant_2022} for a two-class Gaussian example).

\begin{figure}[t]
\begin{center}
  \centerline{\includegraphics[width=0.75\columnwidth]{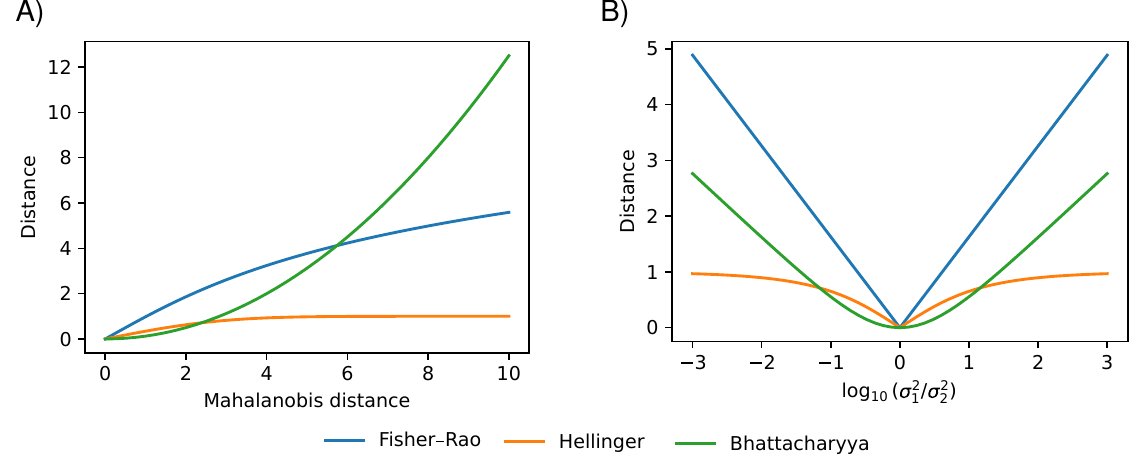}}
\caption{
  Growth in Fisher-Rao, Hellinger, and Bhattacharyya distances for
  special cases with closed-form expressions.
\textbf{A)} Distances between Gaussians with identical covariance but different means
(the Mahalanobis distance is proportional to the distance between the means along a line).
\textbf{B)} Distance between Gaussians in 1D with identical means but different variances.
}
\label{fig:distances_comp}
\end{center}
\end{figure}

Unlike the Fisher-Rao distance, $d_B$, and $d_H$ are not geodesic distances along
a statistical manifold. However, as mentioned,
an appealing property of $d_B$ and $d_H$ for dimensionality
reduction is that they provide bounds on the Bayes classification error between
two classes \citep{fukunaga_introduction_1990}. On the other hand, the
Fisher-Rao distance on the manifold of Gaussian distributions does not have a
direct link to Bayes error. But while this seemingly favors $d_B$ and $d_H$
as learning objectives, the relation to Bayes
error in the two-class case does not always translate well to the
multiclass case \citep{loog_multiclass_2001,thangavelu_multiclass_2008}.
For the multiclass case, the objective function needs to balance
trade-offs between increases in different pairwise distances.
Different distances can place different relative weights on the
class pairs, based on how the distance grows as 
two classes become further apart (Figure~\ref{fig:distances_comp}).
This can lead two different distances to favor different solutions,
even if they are equivalent for the two-class case.
For example, Figure~\ref{fig:distances_comp}A shows
that two pairs of classes with Mahalanobis distances
of 1 and 8, contribute to the Bhattacharyya, Fisher-Rao, and Hellinger
distance objectives with relative weights of 1:64, 1:5, and 1:3 respectively.
The consequences of this are illustrated by a toy
example in Appendix~\ref{apd:distances_equal_cov}, where the Fisher-Rao
distance displays higher robustness to outliers than the Bhattacharyya distance.
Therefore, for the multiclass problem it is largely an empirical question
which objective is better for a given dataset and a given goal.

\subsection{Non-parametric methods based on pairwise sample distances}

Metric learning is another popular family of methods with the
goal of maximizing the distance between data points from different classes,
while minimizing the distance between data points from the same class.
These methods do not assume parametric distributions.

A classic example is Local Fisher Discriminant Analysis
(LFDA) \citep{sugiyama_dimensionality_2007}, which
aims to maximize the ratio of between-class and within-class
scatters like LDA, but it uses local scatter matrices obtained by
weighting pairs of data points by their proximity. This method is useful
when the classes are multimodal, because it does not
force different modes of the same class to be close together, as LDA does.
Another popular method that supports high classification accuracy
is Large Margin Nearest Neighbors (LMNN) \citep{weinberger_distance_2009}.
This method learns a subspace that maximizes the performance
of the kNN classifier by bringing the $k$ nearest neighbors of the
same class closer, while trying to push different classes
apart by a large margin.

Another interesting method with a geometric motivation
is Wasserstein Discriminant Analysis (WDA) \citep{flamary_wasserstein_2018}.
WDA tries to maximize the ratio of between-class and within-class
spread, but it replaces the scatter matrices with regularized Wasserstein
distances between class-conditional empirical distributions.
A regularization hyperparameter controls the trade-off between global and local
structure, with LDA being the special case when the regularization approaches 0.
Unlike the Fisher-Rao distance, the Wasserstein distance is not directly linked
to local discriminability, and it is not invariant to invertible
linear transformations.

This class of methods can capture complex patterns in the data, because
they do not make distributional assumptions. However, they are often
computationally expensive and scale poorly with the number of dimensions
and data samples.

\subsection{Non-parametric methods based on statistical dependence with labels}

Another family of supervised dimensionality reduction methods is built
on the idea of maximizing statistical dependence between the low-dimensional
representation of the data $\mathbf{z}$ and the labels $y$.
Supervised Principal Component Analysis (SPCA) \citep{barshan_supervised_2011} is a
popular representative example. It finds the principal components
that have maximal dependence on the response variable as measured using the
Hilbert-Schmidt independence criterion (HSIC).

%%%%%%%%%%%%%%%%%%%%%%%%%%%%%%%%%%%%%%%%%%%%%%%%%%%%%%%%
%%%%%%%%%%%%%%%%%%%%%%%%%%%%%%%%%%%%%%%%%%%%%%%%%%%%%%%%
%%%%%%%%%%%%%%%%%%%%%%%%%%%%%%%%%%%%%%%%%%%%%%%%%%%%%%%%

\section{Methods}
\label{sec:learning}

\paragraph{Optimization.}
Filters $\F$ were optimized by maximizing Equation~\ref{eq:loss1} using
L-BFGS, until the change in the objective was $\leq 10^{-6}$.
The columns of $\F$ were initialized to random unit vectors
and constrained to have unit norm by parametrizing them in terms
of an unconstrained variable
\citep{lezcano_casado_trivializations_2019}\footnote{Using an orthogonal
constraint did not improve performance or training time.
Additionally, orthogonalization can be performed after learning,
without affecting QDA accuracy.}.
The class-conditional statistics of $\mathbf{z}$ were computed from
the means and covariances of $\mathbf{x}$, i.e.
$\mub_i = \F^T\gamb_i$ and $\Sigmab_i = \F^T\Phib_i\F$.
The optimization procedure is summarized in Algorithm~\ref{alg:sqfa}.

Optimization is non-convex, but different random initializations
generally achieved very similar performance (Appendix~\ref{apd:variability}).
This suggests that SQFA variants are robust to the random initialization seed,
and that random initialization is usually sufficient, although
it might be beneficial to run multiple seeds and select the best solution.

\paragraph{Regularization.}
To prevent rank-deficient or ill-conditioned covariances, a regularization
term was added as $\Sigmab_i = \F^T \Phib_i \F + \mathbf{I}_m\sigma^2$,
where $\mathbf{I}_m$ is the identity matrix, and
$\sigma^2$ is the regularization parameter.
For the digit and neural recording datasets, $\sigma^2$ was selected via
grid-search using a validation set.
Performance was robust across a wide range of $\sigma^2$ values
(Appendix~\ref{apd:regularization_sensitivity}).
For the speed-estimation dataset, we set $\sigma^2=0.001$
to match the biologically-informed value of the
original work \citep{burge_optimal_2015}.

\input{algorithm.tex}

\paragraph{Complexity and scalability.} Each optimization step has complexity
$O(m^3 c^2 + cmn^2)$, where $c$ is the number of classes,
$m$ is the number of filters, and $n$ is the dimensionality of the data
(Appendix~\ref{apd:complexity}). The most expensive computations depend
on the feature space dimension $m$, which is typically small.
A scaling analysis on synthetic data showed that SQFA can learn up to 500 filters,
and handle data with up to 20,000 dimensions
on a consumer laptop over a few minutes (Appendix~\ref{apd:scaling}).
SQFA was among the fastest methods to train in our analyses, usually
converging in a few seconds (Appendix~\ref{apd:training_times}).

\paragraph{Invariance.} The Fisher-Rao distance and the Calvo-Oller bound
are invariant to invertible linear transformations. This implies
that learned filters are unique only up to the subspace they span
(Appendix~\ref{apd:regularization}).
When regularization is used, however, the invariance no longer
holds. This has some consequences for learning. First, in the absence of
regularization the unit norm constraint does not affect the objective, but in
the presence of regularization the constraint is needed to prevent the norm
of the filters from growing indefinitely. Second, in the absence of
regularization the learned filters vary considerably across initializations,
but they become consistent when regularization is used. See 
Appendix~\ref{apd:regularization} for further discussion.

We also note that the filters are not rank-ordered by their usefulness
for classification. To obtain filters rank-ordered by usefulness,
the filters can be learned sequentially in pairs:
two filters are learned first, then this pair can be fixed
and two more can be learned, and so on. The resulting
filter pairs are ordered by how well they separate the classes.

\paragraph{Evaluation.}
To evaluate the usefulness of SQFA for dimensionality reduction,
we compared its performance to some commonly used
linear dimensionality reduction techniques:
PCA, LDA, SPCA \citep{barshan_supervised_2011},
LFDA \citep{sugiyama_dimensionality_2007}, LMNN \citep{weinberger_distance_2009},
and WDA \citep{flamary_wasserstein_2018}.

Additionally, to evaluate how the Fisher-Rao
distance compares to other measures of class dissimilarity for
dimensionality reduction, we used a set of SQFA variants using the same
optimization procedure to maximize different dissimilarity measures. For this
we used the Bhattacharyya distance (SQFA-B), the
Hellinger distance (SQFA-H), the Jeffreys divergence (SQFA-J), and the
Wasserstein distance (SQFA-W)\footnote{For SQFA-W, we constrained the filters to be
orthonormal. This was necessary to prevent SQFA-W filters to
become rank-deficient.}.
Note that SQFA-B is analogous to some popular methods described in the
Related Work \citep{choi_feature_2003,duin_linear_2004},
so it also serves as a benchmark of the literature.

We evaluate the different dimensionality reduction methods using
QDA classification accuracy in the low-dimensional
feature space. We use QDA because
it is the optimal classifier under Gaussian assumptions. Since
SQFA assumes Gaussian class-conditional distributions, it can be seen
as optimizing QDA performance, and
QDA accuracy is a natural way to evaluate SQFA features.
To analyze the more general usefulness of SQFA features, we also
evaluate performance using a kNN classifier in Appendix~\ref{apd:knn}. 
Remarkably, SQFA features support high accuracy even for kNN, and
the conclusions are similar with both evaluation methods.
%(see Appendix~\ref{apd:gaussianity} for a discussion of
%non-Gaussianity). 
Unless otherwise specified, each SQFA variant was trained with 10
different random initializations, and the results show the
median across initializations. 
Like for SQFA, we performed a grid search to select the best regularization
hyperparameters for LDA (covariance shrinkage parameter), LFDA
(number of neighbors), and WDA (regularization parameter).

For LMNN and LFDA, we used the implementations from the Python
package \texttt{metric-learn} \citep{vazelhes_metric-learn_2020}. For WDA, we used
a custom PyTorch implementation following the
Python package \texttt{POT} \citep{flamary_pot_2021}\footnote{The methods
LMNN, LFDA and WDA are very computationally expensive,
so we used a reduced dataset for training them.
For LMNN and WDA we used 500 samples per class. For all three methods,
we reduced the dimensionality down to 200 dimensions using PCA as a
preprocessing step. SQFA performance was similar when using this same
reduced dataset, so our conclusions are not affected by this analysis choice.}.
For PCA, LDA, and QDA we use the package \texttt{scikit-learn} 
\citep{pedregosa_scikit-learn_2011}.

\paragraph{Code.} A Python package implementing SQFA is made available at
\url{https://github.com/dherrera1911/sqfa}.
The code used in our analyses is available
at \url{https://github.com/dherrera1911/sqfa_analyses}.

\begin{figure}[t]
\begin{center}
  \centerline{\includegraphics[width=\columnwidth]{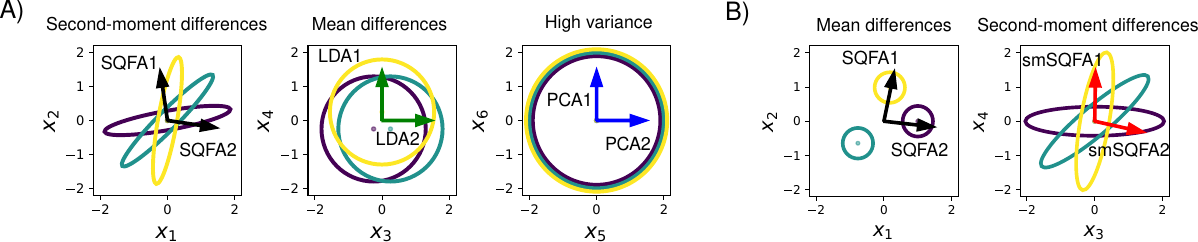}}
\caption{
\textbf{A)} SQFA vs. LDA vs. PCA.
Each of the three panels depicts two dimensions of
a 6D data space. 
Ellipses show the conditional probability distributions of the
data vector $\mathbf{x}$ for three classes
(colors) in a 6D toy dataset.
The classes are separated by different statistical properties.
Classes are distinguished by large differences in the covariances
(dimensions 1-2), small differences in the means (dimensions 3-4),
or neither (dimensions 5-6). Two filters were learned with each of
SQFA, LDA, and PCA. The learned filters are
shown as arrows in the data space, indicating the axis
of the data selected by each filter.
SQFA prefers the most discriminative subspace.
\textbf{B)} SQFA vs. smSQFA. 
Each of the two panels depicts two dimensions of
a 6D data space. 
Ellipses show the conditional probability
distributions of the data vector $\mathbf{x}$ for three classes
(colors) in a 4D toy dataset. Classes are distinguished by large
differences
in the means (dimensions 1-2), and by large differences in the
covariances (dimensions 3-4). We learned two
filters with SQFA and smSQFA, shown as arrows in the
data space. The SQFA filters select for
the most discriminative subspace (dimensions 1-2).
}
\label{fig:toy1}
\end{center}
\end{figure}

\section{Results}

\subsection{Toy problem: SQFA vs. LDA vs. PCA}

First, we illustrate the differences between SQFA and the
canonical dimensionality reduction methods, LDA and PCA.
For this, we designed a toy dataset with a six-dimensional
variable $\mathbf{x}$ and three classes.
The six-dimensional space contains three different 2D subspaces,
each represented by a panel in Figure~\ref{fig:toy1}A.
The statistics of the dataset are built such that each subspace is preferred
by one of the three techniques, i.e. SQFA, LDA, or PCA. With each of
the models we learned two filters on the six-dimensional dataset.
The filters learned by each model are shown as arrows in the data space.

Dimensions 1-2 (Figure~\ref{fig:toy1}A, left) have no differences
between the class means, but have very different--and hence
highly discriminative--class-specific covariances.
This subspace is selected for by SQFA (black arrows), because it
produces the largest Fisher-Rao distances.
Dimensions 3-4 (Figure~\ref{fig:toy1}A, center) contain slight
differences between the class means, but these are not very
discriminative. This subspace is selected for by LDA (green arrows)
because it is the only one with differences in the means.
Dimensions 5-6 (Figure~\ref{fig:toy1}A, right) contain
large covariances, but the class-specific distributions are
identical. Because this subspace contains the largest
overall variances, it is favored by PCA.
Filters that are learned with each of the three
methods--SQFA, LDA, and PCA--select for the expected subspace.

The previous toy problem shows that SQFA can capture the differences
between class-specific covariances that support classification.
However, SQFA can also capture differences in the class means, and
it can flexibly select one or the other depending on
which one is most informative. To illustrate this, we designed a
second toy dataset with three classes and a four-dimensional
variable $\mathbf{x}$, composed of two 2D subspaces
(like the previous example).

Dimensions 1-2 (Figure~\ref{fig:toy1}B, left) contain large differences
in the class-specific means, supporting strong discrimination. Dimensions 3-4
(Figure~\ref{fig:toy1}B, right) have different class-specific
covariances but identical means, supporting weaker discrimination
than dimensions 1-2. We designed the classes such that their
second-moment matrices (i.e. $\Psib_i$) are more
different for dimensions 3-4 than for dimensions 1-2.
We trained both SQFA and smSQFA on the full four-dimensional
dataset, and show the learned filters as arrows in the data space.

Because SQFA can select for either first-order or second-order class
differences, its filters select for dimensions 1-2, where the
classes are more separated (black arrows). On the other hand,
smSQFA is only sensitive to class-specific second-moment matrices,
so its filters select for dimensions 3-4. This shows that SQFA can flexibly
select differences in either the means or the covariances.

It is worth noting that even in the case where the classes
have identical covariance, SQFA can learn features that are different from those
learned by LDA, because both methods maximize different pairwise
distances, so they put different relative weights on the class pairs
(see Section~\ref{sec:related_work_parametric}). A toy
example illustrating this is shown in the Appendix~\ref{apd:distances_equal_cov}.

%%%%%%%%%%%%%%%%%%%%%%%%%%%%%%%%%%%%%%%%%%%%%%%%%%%%%%%%%

\subsection{SQFA for digit classification with poor first-order information}

% table_csvs/svhn_n_filters_qda_review_wide_median.csv
\begin{table}[b]
  \centering
  \small
  \caption{SVHN median QDA classification accuracy (\%). Highest two accuracies are shown in bold.}
  \label{tab:svhn_n_filters_qda}
  \setlength{\tabcolsep}{4pt}
  \begin{tabular}{rcccccccccccc}
    \toprule
    $m$ & SQFA & smSQFA & SQFA-H & SQFA-B & SQFA-W & SQFA-J & LDA & SPCA & LFDA & WDA & LMNN & PCA \\
    \midrule
    2 & \textbf{39.8} & 36.7 & \textbf{39.5} & 36.5 & 33.2 & 36.5 & 23.6 & 19.6 & 22.9 & 22.3 & 19.5 & 19.6 \\
    4 & \textbf{56.4} & \textbf{56.4} & \textbf{56.9} & 55.9 & 48.8 & 55.6 & 27.4 & 22.8 & 30.6 & 39.9 & 25.9 & 24.5 \\
    8 & \textbf{68.1} & \textbf{67.5} & \textbf{68.1} & 66.3 & 58.5 & 67.3 & 34.8 & 37.5 & 33.5 & 57.1 & 49.9 & 35.6 \\
    16 & 74.7 & 74.6 & \textbf{74.8} & 74.4 & 71.8 & \textbf{75.2} & -- & 50.4 & 34.4 & 66.1 & 67.5 & 58.4 \\
    \bottomrule
  \end{tabular}
\end{table}

To examine how SQFA performs on high-dimensional real-world data,
we tested it using the grayscale Street View House Numbers (SVHN) dataset,
composed of 1024-dimensional images. The many sources of variation in the images
(e.g.\ digits with mixed contrast polarity, variation in background intensity),
make first-order differences between classes a poor signal for discriminating
between the digits, and make second-order statistics important for classification performance. For each filter-learning method, we learned 2, 4, 8, and 16
filters ($m$). We then evaluated the QDA classification accuracy in each
low-dimensional feature space (note that LDA cannot have
$m > c-1$, where the number of classes $c$ is 10 for this dataset).

Notably, all variants of SQFA achieved higher QDA accuracy than the remaining methods
for all values of $m$ (Table~\ref{tab:svhn_n_filters_qda}). 
SQFA and SQFA-H were in general the best performing methods,
being outperformed by SQFA-J only for $m=16$. It is noteworthy
that SQFA and SQFA-H both consistently
outperform SQFA-B, which is the more commonly used dissimilarity measure in the
dimensionality reduction literature
\citep{choi_feature_2003,duin_linear_2004}. smSQFA performed similarly to
SQFA, in line with the class-conditional means containing little information
for this dataset. Remarkably, this pattern of results was almost identical when
evaluated with kNN accuracy, with SQFA and SQFA-H being the best
performing methods (Appendix~\ref{apd:knn}).

%\begin{figure}[t]
%\begin{center}
%\centerline{\includegraphics[width=\columnwidth]{figures/svhn.pdf}}
%\caption{SQFA extracts useful features using class-conditional second-order
%statistics. \textbf{A)} Example images
%from SVHN. \textbf{B)} QDA accuracy using the features learned by the
%different methods. For SQFA variants, the median and interquartile range
%of 20 different initializations are shown.
%\textbf{C)} Filters learned by the methods.}
%\label{fig:digits}
%\end{center}
%\end{figure}

\subsection{SQFA for digit classification with useful first-order information}
\label{sec:mnist}

To examine how SQFA performs when both first- and second-order statistics
are informative, we compared the same methods using the MNIST dataset.
Because MNIST digits are white digits on a black
background, the class-conditional means are quite different,
making them useful for classification. The same number of filters were learned for
each method as for the SVHN dataset.

SQFA-H achieved the highest QDA accuracy for all values of $m$
(Table~\ref{tab:mnist_n_filters_qda}). 
For most values of $m$, SQFA-B had the second highest performance.
SQFA was competitive, performing better than many
other methods (it placed either 3rd or 5th across values of $m$, out of 12 methods).
Thus, when the class-conditional means are informative, maximizing
the Fisher-Rao distance can be competitive, but maximizing the Hellinger distance
can be the best option. The conclusions are the same when using kNN classifier accuracy
(Appendix~\ref{apd:knn}).

% table_csvs/mnist_n_filters_qda_review_wide_median.csv
\begin{table}[t]
  \centering
  \small
  \setlength{\tabcolsep}{4pt}
  \caption{MNIST median QDA classification accuracy (\%). Highest two accuracies are shown in bold.}
  \label{tab:mnist_n_filters_qda}
  \begin{tabular}{rcccccccccccc}
    \toprule
    $m$ & SQFA & smSQFA & SQFA-H & SQFA-B & SQFA-W & SQFA-J & LDA & SPCA & LFDA & WDA & LMNN & PCA \\
    \midrule
    2 & 59.7 & \textbf{62.4} & \textbf{66.6} & 56.0 & 52.6 & 50.1 & 56.6 & 48.4 & 59.2 & 55.0 & 52.6 & 46.1 \\
    4 & 80.0 & 76.6 & \textbf{86.6} & \textbf{82.5} & 74.6 & 76.8 & \textbf{82.5} & 72.9 & 82.2 & 67.8 & 73.5 & 63.1 \\
    8 & 89.7 & 87.8 & \textbf{93.2} & \textbf{90.8} & 88.7 & 88.7 & 90.1 & 88.5 & 89.4 & 86.8 & 90.2 & 86.5 \\
    16 & 94.2 & 94.0 & \textbf{95.4} & \textbf{94.4} & 94.1 & 94.0 & -- & 92.5 & 90.5 & 93.8 & 94.2 & 93.7 \\
    \bottomrule
  \end{tabular}
\end{table}

%\begin{figure}[ht]
%\begin{center}
%\centerline{\includegraphics[width=\columnwidth]{figures/mnist.pdf}}
%\caption{SQFA can exploit class-conditional first- and second-order information.
%\textbf{A)} Example MNIST images.
%\textbf{B)} QDA accuracy using the features
%learned by the different methods.
%For SQFA variants, the median and interquartile range
%of 20 initializations are shown.
%\textbf{C)} Filters learned by the methods.
%}
%\label{fig:mnist}
%\end{center}
%\end{figure}

\subsection{Naturalistic speed-estimation task}

Next, we examine how SQFA features perform on a naturalistic speed-estimation
dataset used to investigate neural computations
\citep{burge_optimal_2015,chin_predicting_2020,herrera-esposito_optimal_2024}.
This video-based dataset is interesting for several reasons.
First, finding features that are useful for solving visual
tasks is essential for sensory-perceptual neuroscience \citep{burge_image-computable_2020},
and a potential application for SQFA. 
Second, the class-conditional distributions are well approximated by zero-mean
Gaussians, which means that the assumptions of smSQFA are approximately
satisfied. Thus, the results with smSQFA for this dataset are a good
approximation to the filters obtained by maximizing the true Fisher-Rao
distance, as opposed to the Calvo-Oller bound used by SQFA.
Third, a method called AMA-Gauss which directly maximizes the performance of
a Bayesian Gaussian decoder is reported to perform well for this
dataset \citep{jaini_linking_2017}, providing a principled benchmark for
comparison (see Appendix~\ref{apd:speed}). 

% table_csvs/motion_n_filters_qda_review_wide_median.csv
\begin{table}[b]
  \centering
  \small
  \setlength{\tabcolsep}{4pt}
  \caption{Speed estimation median QDA classification accuracy (\%). Highest two accuracies are shown in bold.}
  \label{tab:motion_n_filters_qda}
  \begin{tabular}{rccccccccccccc}
    \toprule
    $m$ & SQFA & smSQFA & SQFA-H & SQFA-B & SQFA-W & SQFA-J & AMA & LDA & SPCA & LFDA & WDA & LMNN & PCA \\
    \midrule
    2 & 58.1 & \textbf{58.8} & 56.0 & 58.1 & 31.3 & 58.7 & \textbf{61.9} & 5.4 & 9.5 & 4.9 & 9.4 & 4.7 & 23.9 \\
    4 & 68.0 & \textbf{68.3} & 67.3 & 66.4 & 45.4 & 65.8 & \textbf{74.1} & 11.2 & 31.3 & 25.3 & 18.0 & 15.3 & 32.8 \\
    6 & 80.7 & 79.6 & \textbf{83.9} & 78.3 & 61.5 & 68.5 & \textbf{85.9} & 20.9 & 40.3 & 11.5 & 30.3 & 35.3 & 48.7 \\
    8 & 89.3 & \textbf{89.4} & 89.2 & 84.9 & 80.8 & 69.9 & \textbf{91.9} & 28.1 & 59.8 & 52.8 & 38.6 & 51.0 & 75.1 \\
    \bottomrule
  \end{tabular}
\end{table}

\begin{figure}[t]
\begin{center}
\centerline{\includegraphics[width=\columnwidth]{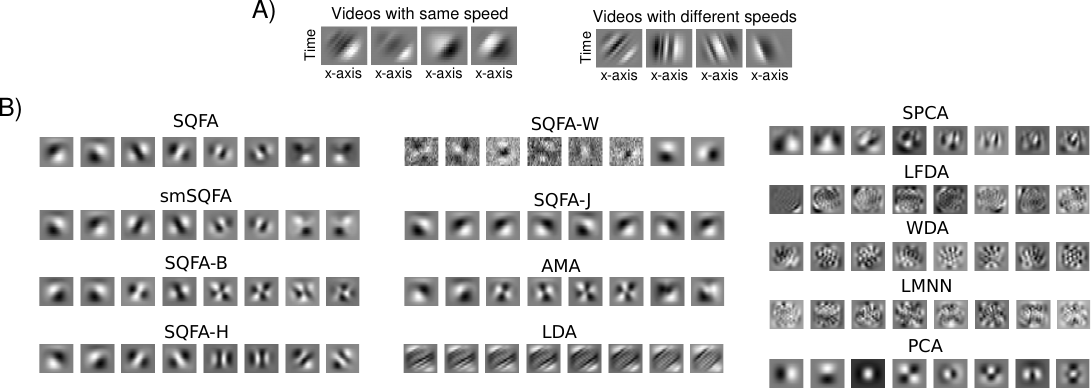}}
\caption{SQFA learns motion-sensitive features similar to those found in
biological neural systems.
\textbf{A)} 4 example videos with the same speed (left)
and 4 example videos with different speeds (right).
Each video is shown as a 2D space-time plot where the vertical axis is time
and the horizontal axis is space.
\textbf{C)} Filters learned by the methods (each image shows
a 2D space-time plot).
}
\label{fig:speed}
\end{center}
\end{figure}

Each video consists of 30 horizontal pixels and 15 frames.
The vertical dimension was averaged out, hence the videos can be represented
as 2D space-time plots (Figure~\ref{fig:speed}A). Each video shows a naturally
textured surface moving with one of 41 different speeds (i.e.\ classes).
We learned up to 8 filters with each method, following the original
work \citep{burge_optimal_2015}.

As expected, AMA-Gauss performed best, since it directly optimizes
the decodability of speed under Gaussian assumptions
(Table~\ref{tab:motion_n_filters_qda}). smSQFA is the
second-best method for most values of $m$, and SQFA-H was also a
top-ranking method across values of $m$. SQFA-B achieved high
performance, but not as good as smSQFA and SQFA-H. 
Similar conclusions are obtained when using the kNN classifier instead of QDA
(Appendix~\ref{apd:knn}, although see the accompanying discussion).
We note that SQFA variants do not achieve as high QDA
accuracy as AMA-Gauss filters because of two reasons:
the dissimilarity measures used by SQFA variants are not perfect measures of
decodability, and the average of pairwise discriminabilities 
is not guaranteed to maximize multiclass discriminability
\citep{loog_multiclass_2001,thangavelu_multiclass_2008}
(see Section~\ref{apd:distances_equal_cov}).
The fact that SQFA and SQFA-H perform so close to AMA-Gauss
highlights the utility of Fisher-Rao and Hellinger distances as
dimensionality reduction objectives.

Finally, we visualized the features learned by the different methods.
To obtain more interpretable filters, we used pairwise filter learning
when visualizing the SQFA variants (see Section~\ref{sec:learning}).
The best performing SQFA variants (SQFA, smSQFA, SQFA-H and SQFA-B),
as well as AMA-Gauss, learned filters that are similar to typical motion-sensitive
receptive fields in visual cortex, selecting for a range of spatio-temporal frequencies
\citep{movshon_spatial_1978,rust_spatiotemporal_2005,priebe_tuning_2006}.
In contrast, filters learned by all other methods either lack clear motion
sensitivity, or do not cover a range of spatio-temporal frequencies
(Figure~\ref{fig:speed}C). This shows that SQFA can learn features that
are both useful for classification and interpretable.

\subsection{Neural data analysis}

Next, to show the applicability of SQFA to real-world experimental data,
we tested SQFA on an open dataset of neural spike recordings
\citep{zandvakili_coordinated_2015,amin_zandvakili_paired_2019}.
SQFA can be useful to analyze neural data for different reasons.
First, because stimulus-dependent (i.e. class-specific) response covariances
are considered important for neural
coding \citep{moreno-bote_information-limiting_2014,kohn_correlations_2016},
and finding the informative low-dimensional subspaces that
account for stimulus-dependent covariances is an important problem in neuroscience.
Second, there is interest in studying the geometry of neural representations
\citep{kriegeskorte_neural_2021,wang_geometry_2021,chung_neural_2021},
for which information geometry can provide useful tools \citep{kriegeskorte_neural_2021}.
It is also worth noting that spiking neural responses present an interesting
challenge for SQFA, because they are noisy and highly non-Gaussian.

The responses were recorded from primary visual cortex of macaque
monkeys shown drifting gratings of 8 different orientations (classes).
The dataset includes 5 recording sessions from 3 animals, each session with
400 trials per class, and between 70 and 142 recorded neurons.
We removed outlier trials (1.2\%) and neurons (3.3\%) (see Appendix~\ref{apd:neural}).
Because this dataset consists of different sessions that cannot
be used to learn a single set of features, we used a modified evaluation procedure.
For each session we performed 5 different random stratified splits of the data
into training, validation, and test sets, resulting in 25 total splits.
For each split we learned the filters and evaluated QDA classification accuracy.
We report the mean QDA accuracy across all splits\footnote{We use the mean for this dataset
because performance across splits is highly correlated within sessions.
The mean better summarizes the performance across sessions than the median,
which reflects only the session with the middle performance. The same conclusion
are obtained when using the median.}.
For more information about the dataset see
\cite{amin_zandvakili_paired_2019}.

% table_csvs/zand_kohn_n_filters_qda_review_wide_mean.csv
\begin{table}[t]
  \centering
  \small
  \setlength{\tabcolsep}{4pt}
  \caption{Neural data mean QDA classification accuracy (\%). Highest two accuracies are shown in bold.}
  \label{tab:zand_kohn_n_filters_qda}
  \begin{tabular}{rcccccccccccc}
    \toprule
    $m$ & SQFA & smSQFA & SQFA-H & SQFA-B & SQFA-W & SQFA-J & LDA & SPCA & LFDA & WDA & LMNN & PCA \\
    \midrule
    2 & 90.2 & 88.1 & \textbf{91.4} & 88.4 & 86.0 & 86.9 & 88.7 & 85.6 & 88.1 & 88.0 & \textbf{91.0} & 68.4 \\
    4 & 94.4 & 94.2 & \textbf{96.1} & 93.2 & 85.4 & 92.1 & 93.8 & 92.5 & 93.6 & 91.7 & \textbf{95.0} & 91.5 \\
    8 & 96.0 & 95.4 & \textbf{96.8} & \textbf{96.5} & 83.1 & 95.7 & -- & 96.4 & 95.9 & 87.2 & \textbf{96.7} & 96.4 \\
    \bottomrule
  \end{tabular}
\end{table}

Similar to previous datasets, SQFA-H had the highest performance for all values of $m$
(Table~\ref{tab:zand_kohn_n_filters_qda}). LMNN had the second highest performance,
and SQFA was the third-best performing method for $m=2$ and $m=4$, again proving
that maximizing the Fisher-Rao distances is a competitive strategy for dimensionality
reduction. Similar conclusions are obtained when using the kNN classifier
(Appendix~\ref{apd:knn}). These results show that SQFA and
its variants can be useful for finding low-dimensional subspaces of neural
data that make the best use of stimulus-dependent covariances for classification.
Additionally, the results show that SQFA can perform well even when the
data are highly non-Gaussian.

\subsection{Maximizing the Calvo-Oller bound leads to largest Fisher-Rao distances}

Lastly, we evaluated the performance of the Calvo-Oller bound as a surrogate
for the Fisher-Rao distance in the learning objective. 
First, we asked whether maximizing the Calvo-Oller bound leads
to the largest Fisher-Rao distance among the different objectives.
For the filters learned in each of the datasets above by LDA, PCA, and each SQFA
variant, we numerically computed the average Fisher-Rao
distance between the class distributions in the learned feature
spaces using the method of \cite{nielsen_simple_2023}, implemented in the package
\texttt{pyBregMan} \citep{nielsen_pybregman_2024}.
For all datasets and values of $m$, maximizing the
Calvo-Oller bound led to the highest average Fisher-Rao distance
in the feature space (Table~\ref{tab:mean_fr_distance}).

\begin{table}[bh]
  \centering
  \caption{Mean Fisher--Rao class distance in feature space for different optimization objectives. Columns indicate the distance maximized by the SQFA variants, or the learning algorithm.}
  \begin{tabular}{rcccccccc}
    \hline
    $m$ & Dataset & Calvo--Oller & Hellinger & Bhattacharyya & Wasserstein & Jeffreys & PCA & LDA \\
    \hline
    2  & SVHN & \textbf{1.42} & 1.20 & 1.35 & 1.20 & 1.36 & 0.12 & 0.37 \\
    4  & SVHN & \textbf{2.31} & 2.24 & 2.29 & 1.87 & 2.27 & 0.55 & 0.92 \\
    8  & SVHN & \textbf{2.51} & \textbf{2.51} & 2.45 & 2.12 & 2.47 & 1.13 & 1.07 \\
    16 & SVHN & \textbf{3.50} & 3.49 & 3.49 & 3.05 & 3.44 & 2.26 & -- \\
    2  & MNIST & \textbf{2.92} & 2.86 & 2.80 & 2.66 & 2.73 & 2.40 & 2.22 \\
    4  & MNIST & \textbf{3.90} & 3.79 & 3.84 & 3.54 & 2.77 & 3.21 & 3.16 \\
    8  & MNIST & \textbf{3.85} & 3.49 & 3.56 & 3.66 & 3.76 & 3.56 & 3.02 \\
    16 & MNIST & \textbf{4.62} & 4.31 & 4.55 & 4.40 & 4.42 & 4.34 & -- \\
    2 & Speed & \textbf{3.34} & \textbf{3.34} & \textbf{3.34} & 2.97 & 3.34 & 2.81 & 1.30 \\
    4 & Speed & \textbf{4.33} & 4.27 & 4.32 & 3.97 & 3.91 & 3.72 & 3.09 \\
    8 & Speed & \textbf{5.36} & 5.19 & 5.27 & 5.09 & 4.51 & 4.86 & 5.26 \\
    2 & Neural data & \textbf{3.81} & 3.78 & 3.75 & 3.47 & 3.78 & 2.95 & 3.79 \\
    4 & Neural data & \textbf{4.83} & 4.56 & 4.68 & 4.00 & 4.81 & 4.55 & 4.56 \\
    8 & Neural data & \textbf{5.42} & 5.18 & 4.91 & 5.30 & 5.40 & 4.98 & -- \\
\hline
\end{tabular}
  \label{tab:mean_fr_distance}
\end{table}

\begin{figure}[t]
\begin{center}
\centerline{\includegraphics[width=\textwidth]{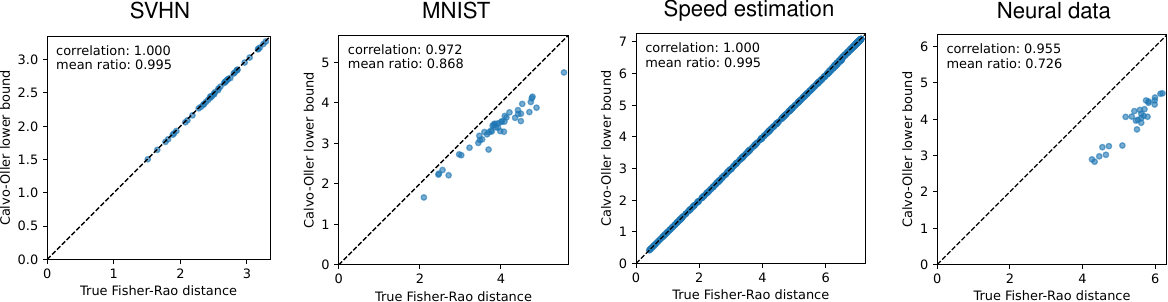}}
\caption{Calvo-Oller bound vs.\@ true Fisher-Rao distance
in real-world datasets. Each panel shows the Calvo-Oller bound
and the Fisher-Rao distance for the class pairs of a given dataset.
From left to right, the SVHN, MNIST and the speed estimation
dataset are shown. Each point represents a different pair of classes.
The dashed line shows the identity line. The correlation and the
mean of the pairwise ratios between the Calvo-Oller bound and the true
Fisher-Rao distance are shown in the top left of each panel.
}
\label{fig:calvo_oller}
\end{center}
\end{figure}

Then, we compared the Calvo-Oller bounds to the
Fisher-Rao distances in the SQFA feature space across all class pairs.
Figure~\ref{fig:calvo_oller} shows the comparison for $m=8$.
The pairwise Calvo-Oller bounds are highly correlated with the
true Fisher-Rao distances in all datasets. Furthermore, the Calvo-Oller
bounds are close to the true Fisher-Rao distance, as indicated by the
mean ratios between the two.
The largest deviation was observed for the neural dataset, which has a
mean ratio of 0.726, indicating that the Calvo-Oller bound was on average
27.4\% smaller than the true Fisher-Rao distance.
The Calvo-Oller bound and the Fisher-Rao distance are almost identical
for the SVHN and speed estimation datasets,
which is expected because these datasets have similar means across classes,
and the Calvo-Oller bound is exact in the equal-mean case
(Appendix~\ref{apd:calvo_oller}).

In sum, these results show that maximizing the Calvo-Oller bound is a
practical way to maximize the Fisher-Rao distance for learning
in real-world datasets.

\section{Discussion}

We have introduced SQFA, a supervised dimensionality reduction
method that maximizes the Fisher-Rao distances
between class-conditional distributions under Gaussian assumptions.
SQFA is a computationally efficient method that learns
features supporting excellent quadratic decodability, i.e.\@ QDA
classification accuracy.
The same optimization procedure can be used to define variants of
SQFA that maximize other dissimilarity measures from statistics
and information theory.

The results exhibit noteworthy patterns.
First, maximizing the Fisher-Rao distance, which is derived in a
geometric framework, leads to features that support similar performance 
as maximizing the Bhattacharyya and Hellinger distances, which are
information-theoretic dissimilarity measures directly
linked to Bayes classification error.
Also, SQFA features often supported higher QDA and kNN accuracy
than popular dimensionality reduction methods such as LDA, SPCA, LFDA,
WDA and LMNN. Therefore, SQFA is a competitive method for
supervised dimensionality reduction that can be useful in combination
with different types of decoders.
Notably, maximizing the Fisher-Rao distance led to much better discriminative
features than maximizing the Wasserstein distance (SQFA-W).
Hence, Riemannian distances between class-conditional distributions that do not
reflect discriminability (e.g. Wasserstein) are not necessarily good objectives
for learning discriminative features.

Second, despite the fact that the Hellinger and Bhattacharyya distances
are equivalent objectives for the two-class case,
SQFA-H consistently outperformed SQFA-B in our analyses.
Moreover, SQFA-H was in general the best performing method
across datasets and number of filters learned, both for QDA and for kNN accuracy. 
Why does maximizing the Hellinger distance lead to better performance than
maximizing the Bhattacharyya distance? One likely explanation relates to
the fact that the sum of pairwise discriminabilities does not perfectly
translate to multiclass discriminability
\citep{loog_multiclass_2001,tao_general_2007,thangavelu_multiclass_2008}.
In particular, the behavior of the multiclass objective depends on how it
weights the discriminabilities of the different pairs of classes.
Because dissimilarity measures can scale differently as distributions
become further apart, the relative contributions of the class pairs 
to the objective in Equation~\ref{eq:loss1} can be different
(see Section~\ref{sec:related_work_parametric}).
For example, the Hellinger distance is bounded between 0 and 1, and classes that
are already well separated will not contribute much to the loss gradient,
so the learning procedure will emphasize separating classes that are close together.
In contrast, the Bhattacharyya distance grows faster as two classes are further apart,
which can favor separating classes that are already well separated,
leading to little improvement in classification performance
(see Appendix~\ref{apd:distances_equal_cov}).
The advantage of the Hellinger distance over the Bhattacharyya distance observed
in our analyses should be of interest for practitioners, because
the Bhattacharyya distance is the most common choice in linear dimensionality reduction.
Notably, the Fisher-Rao distance often grows at a rate between
those of the Bhattacharyya and the Hellinger distances, which might
help explain the good performance of SQFA.

A novel methodological aspect of this work is the use of the Calvo-Oller
bound as a surrogate for the Fisher-Rao distance in learning
\citep{calvo_distance_1990,nielsen_simple_2023}.
The fact that closed-form expressions for the Fisher-Rao distance between
multivariate Gaussians are unavailable has limited its use in practice.
Besides the success of the Calvo-Oller bound at learning discriminative
features, we showed that maximizing the Calvo-Oller bound led to a
higher Fisher-Rao distance than maximizing any other objective,
and that the pairwise Calvo-Oller bounds were highly correlated
with the true pairwise Fisher-Rao distances. This motivates further
explorations of the Calvo-Oller bound as an objective for machine
learning applications.
Additionally, formulas for the Calvo-Oller bound exist for other elliptical
distributions such as multivariate Student-t and Cauchy distributions
\citep{calvo_distance_2002,nielsen_simple_2023}. Future work can use
these formulas to extend SQFA to other non-Gaussian elliptical distributions.

Information geometry is a promising
tool for studying neural representations in neuroscience and machine learning
\citep{kriegeskorte_neural_2021,wang_geometry_2021,
arvanitidis_pulling_2022,duong_representational_2023,feather_discriminating_2024}.
Finding features that maximize Fisher-Rao distances between classes,
or conditions, is a potentially useful tool in this context.
Recent research in neuroscience and psychology suggests
a relation between geodesic distances in perceptual space (using a metric
analogous to the Fisher metric) and similarity judgments
\citep{riemann_ueber_1867,fechner_elemente_1860,bujack_non-riemannian_2022,
zhou_unified_2024,vacher_perceptual_2024,hong_comprehensive_2025}.
SQFA shows that maximizing a geometric objective with similar distances
can be useful for supervised dimensionality reduction.
%One concrete example of a potential application of SQFA in neuroscience
%is to analyze the magnitude and relevance of the condition-dependent
%covariability of neural responses, which is a topic of much debate,
%and for which the Fisher information has been used extensively
%\citep{moreno-bote_information-limiting_2014,kohn_correlations_2016,ding_information_2023}.
%SQFA can find the modes of neural activity where different conditions
%differ most in their first- and second-order neural responses,
%potentially providing insights into the relevance of condition-dependent
%covariability. SQFA can also leverage modern tools from neuroscience
%that estimate the response statistics in conditions where raw data itself is not available
%\citep{nejatbakhsh_estimating_2023,ding_information_2023,maheswaranathan_interpreting_2023}.

SQFA is a first step towards using information geometry for
dimensionality reduction (also see
\cite{carter_information_2009,carter_information-geometric_2011,dwivedi_discriminant_2022}).
Several questions and extensions remain open.
For example, promising directions include extending SQFA to non-Gaussian
elliptical distributions, or even to the non-parametric case,
using the non-parametric Fisher-Rao distance
\citep{srivastava_riemannian_2007}.
Future work should also study under what circumstances
maximizing the Fisher-Rao distance might be preferable to
maximizing other measures of dissimilarity, such as the
Hellinger distance. Finally, it should be straightforward to extend this
framework to nonlinear feature learning, a topic that will
be explored in future work.

\bibliography{main.bib}
\bibliographystyle{tmlr}

\newpage

\appendix

%%%%%%%%%%%%%%%%%%%%%%%%%%%%%%%%%%%%%%%%%%%%%%%%%%%%%%%%%%%%%%%%%%%%%%%%%%%
%%%%%%%%%%%%%%%%%%%%%%%%%%%%%%%%%%%%%%%%%%%%%%%%%%%%%%%%%%%%%%%%%%%%%%%%%%%
%%%%%%%%%%%%%%%%%%%%%%%%%%%%%%%%%%%%%%%%%%%%%%%%%%%%%%%%%%%%%%%%%%%%%%%%%%%

\section{Fisher-Rao distance and discriminability in the zero-mean case}
\label{apd:zero_mean}

\subsection{Generalized eigenvalues reflect quadratic differences between classes}
\label{apd:dist_theory}

As described in the main text, the Fisher-Rao distance between
two zero-mean Gaussian distributions $\mathcal{N}(0,\Psib_i)$ and
$\mathcal{N}(0,\Psib_j)$ is given (up to a factor of $\sqrt{2}$)
by the affine-invariant distance in the manifold of
symmetric positive definite matrices, $\mathrm{SPD(m)}$
\begin{equation}
  \label{eq:ai2}
  d_{FR}(\Psib_i,\Psib_j) =  \sqrt{\frac{1}{2}\sum_{k=1}^m \log^2(\lambda_k)}
\end{equation}
where $\lambda_k$ are the generalized eigenvalues of the
pair of matrices $(\Psib_i,\Psib_j)$.

The generalized eigenvalues $\lambda_k$ and generalized eigenvectors
$\mathbf{v}_k$ of the pair of matrices
$\mathbf{A} \in \mathrm{SPD(m)}$ and
$\mathbf{B} \in \mathrm{SPD(m)}$ are the
solutions to the generalized eigenvalue problem
$\mathbf{A}\mathbf{v}_k = \lambda_k \mathbf{B}\mathbf{v}_k$.
The solution to the problem is given
by the eigenvalues and eigenvectors of
$\mathbf{B}^{-1/2}\mathbf{A}\mathbf{B}^{-1/2}$,
where $\mathbf{B}^{-1/2}$ is the inverse square root of $\mathbf{B}$.
If $\mathbf{A}$ and $\mathbf{B}$ are identical,
$\mathbf{B}^{-1/2}\mathbf{A}\mathbf{B}^{-1/2}$
is the identity matrix, all the eigenvalues are $1$, and
$d_{FR}(\A,\B)=0$.
The farther the $\lambda_k$ are from $1$, the more
different the matrices $\mathbf{A}$ and $\mathbf{B}$ are
(i.e.\ the more different $\mathbf{B}^{-1/2}\mathbf{A}\mathbf{B}^{-1/2}$
is from the identity matrix).

Consider a random variable $\mathbf{z} \in \mathbb{R}^m$ that belongs to
one of two classes $i,j$, with second moment matrices
$\Psib_i = \mathbb{E} \left[ \mathbf{z}\mathbf{z}^T|y=i \right]$ 
and $\Psib_j = \mathbb{E} \left[ \mathbf{z}\mathbf{z}^T|y=j \right]$.
Next, consider a vector $\mathbf{w} \in \mathbb{R}^m$ and the squared
projection of $\mathbf{z}$ onto $\mathbf{w}$, $(\mathbf{w}^T\mathbf{z})^2$.
The following ratio relates to how different the squared projections
are for the two classes, which is a useful proxy for quadratic
discriminability:
\begin{equation}
  \label{eq:ratio2}
  R(\mathbf{w}) = \frac{\mathbb{E}[(\mathbf{w}^T\mathbf{z})^2|y=i]}{\mathbb{E}[(\mathbf{w}^T\mathbf{z})^2|y=j]} = \frac{\mathbf{w}^T\Psib_i\mathbf{w}}{\mathbf{w}^T\Psib_j\mathbf{w}}
\end{equation}
The local extrema of the ratio $R(\mathbf{w})$
are obtained at the generalized eigenvectors
$\mathbf{w} = \mathbf{v}_k$ of $(\Psib_i,\Psib_j)$, where
the ratio $R(\mathbf{v}_k) = \lambda_k$ \citep{fukunaga_introduction_1990}.

The more different $\lambda_k$ is from $1$, the more
different are the expected squared projections
$(\mathbf{v}^T\mathbf{z})^2$ for the two classes.
The magnitude of $\log^2 \lambda_k$ indicates
how different the ratio in Equation~\ref{eq:ratio2}
is from $1$, in proportional terms.
The set of $\mathbf{v}_k$'s spans the space
of $\mathbf{z}$, so $d_{FR}(\Psib_i,\Psib_j)$
summarizes the quadratic differences
between the classes $i,j$ along all directions in the feature
space, thus relating to the quadratic discriminability of the classes.

Of course, the quadratic discriminability between the classes
depends on factors other than the ratio of the expected values
of the squared projections, and larger differences in this ratio
do not strictly indicate higher discriminability. However,
empirical studies show that the generalized eigenvalues tend to
be a good indicator of quadratic discriminability in real
world datasets \citep{karampatziakis_discriminative_2014}. 

%%%%%%%%%%%%%%%%%%%%%%%%%%%%%%%%%%%%%%%%%%%%%%%%%%%%%%%%%%%%%%%%%%%%%%%%%%%

\subsection{Fisher-Rao distance and Bayes error}
\label{apd:bayes_error}

Here, we explicitly compare the Fisher-Rao distance to
the Bayes error for zero-mean Gaussians in the 1D and 2D cases.

Two symmetric positive definite (SPD) matrices
$\Psib_i$ and $\Psib_j$ can be simultaneously diagonalized
by a linear transformation of the data space. Given the
affine-invariance of the Fisher-Rao distance and of the Bayes
error, we can reduce their analysis 
to the analysis of Gaussians of the form $\mathcal{N}(0,\mathbf{D})$ and
$\mathcal{N}(0,\mathbf{I})$, where $\mathbf{D}$ is a diagonal matrix and
$\mathbf{I}$ is the identity. Then, the Fisher-Rao and the Bayes error are
a function of the diagonal elements of $\mathbf{D}$.
The diagonal elements of $\mathbf{D}$ are the generalized eigenvalues of 
$(\mathbf{D},\mathbf{I})$, so the Fisher-Rao distance between
$\mathcal{N}(0,\mathbf{D})$ and $\mathcal{N}(0,\mathbf{I})$ is given by
\begin{equation}
  \label{eq:fr_diag}
d_{FR}(\mathbf{D},\mathbf{I}) = \sqrt{\frac{1}{2}\sum_{k=1}^m \log^2(\mathbf{D}_{kk})}
\end{equation}

We computed the Bayes error ($e_B$) for each value of $\mathbf{D}$
by simulating 100,000 samples from each $\mathcal{N}(0,\mathbf{D})$ and
$\mathcal{N}(0,\mathbf{I})$ distribution, and obtaining the
error rate of the optimal probabilistic classifier.
We compute the Bayes accuracy ($a_B$) as $a_B = 1 - e_B$.
Because the accuracy is bounded to be between 0 and 1, we
also computed the log-odds of correctly classifying a sample,
given by $\log\left(\frac{a_B}{e_B}\right)$.

\begin{figure*}[t]
\begin{center}
\centerline{\includegraphics[width=0.7\textwidth]{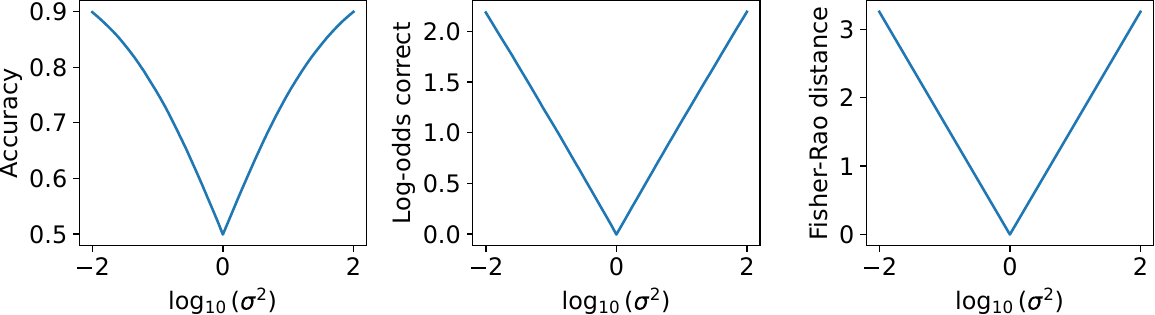}}
\caption{Distances and Bayes error for 1D Gaussian distributions.
From left to right, the three panels show, as a function
of $\log_{10}(\sigma^2)$, accuracy of the Bayes classifier,
the log-odds ratio of correct classification, and
the Fisher-Rao distance.
}
\label{fig:bayes1D}
\end{center}
\end{figure*}

\paragraph{Zero-mean Gaussian, 1D case.} In the 1D case,
the matrix $\mathbf{D}$ is a single positive number $\sigma^2$.
Figure~\ref{fig:bayes1D} shows the
Bayes error, the log-odds of correct classification, and the
Fisher-Rao distance as a function of $\log \sigma^2$.

The Fisher-Rao distance grows linearly with
$|\log_{10}\sigma^2|$ in the 1D case.
Interestingly, the Bayes accuracy is approximately
linear for small values of $\sigma^2$, although it begins to saturate
for larger values of $\sigma^2$ (since it is bounded by 1).
The log-odds of a correct classification also looks like a linear
function of $|\log_{10}\sigma^2|$.
This suggests that the Fisher-Rao distance is a good
proxy for classification performance in the zero-mean 1D Gaussian case.

\begin{figure*}[t]
\begin{center}
\centerline{\includegraphics[width=0.8\textwidth]{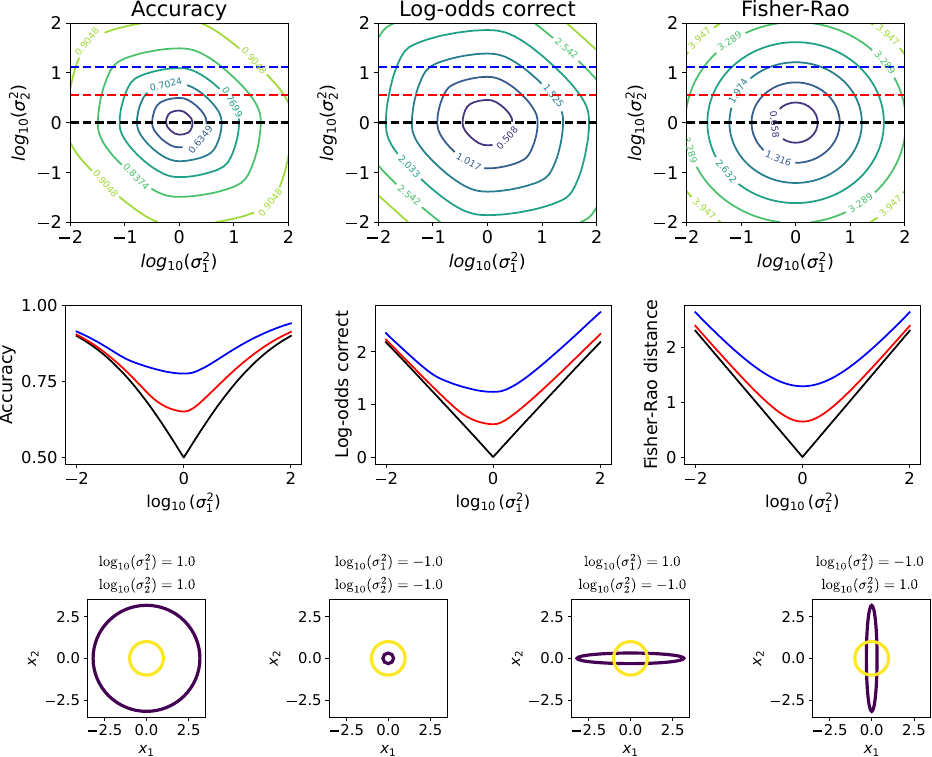}}
\caption{Distances and Bayes error for 2D Gaussian distributions.
From left to right, the panels show for two Gaussian classes,
the accuracy of the Bayes classifier,
the log-odds of a correct classification, and the Fisher-Rao distance.
\textbf{Top}. Contour plots of the quantities as a function
of $\log_{10}(\sigma_1^2)$ and $\log_{10}(\sigma_2^2)$.
The values for the contour lines are shown as a color map.
The horizontal dashed lines indicate the 1D slices
that are shown in the middle row.
\textbf{Middle}. The same quantities as the top row,
shown as a function of
$\log_{10}(\sigma_1^2)$, for fixed values of
$\log_{10}(\sigma_2^2)$. The values used are 0.0 (black),
0.56 (red), and 1.12 (blue).
\textbf{Bottom}. Examples of two zero-mean distributions
for different values of $\log_{10}\sigma_1^2$ and
$\log_{10}\sigma_2^2$.
}
\label{fig:bayes2D}
\end{center}
\end{figure*}

\paragraph{Zero-mean Gaussian, 2D case.} Next we analyze the 2D case.
Here, there are two parameters, $\sigma_1^2$ and $\sigma_2^2$.
The top row of Figure~\ref{fig:bayes2D} shows the level sets of
accuracy, the log-odds correct, and the Fisher-Rao distance
as a function of $\log_{10}\sigma_1^2$ and $\log_{10}\sigma_2^2$.
The level sets of the Fisher-Rao distance have a different shape than
the level sets of the accuracy and log-odds correct.
The contours of the accuracy and log-odds correct have
more circular shapes close to the origin, but acquire a
tilted hexagonal shape farther from the origin. 
As expected from Equation~\ref{eq:ai2}, the Fisher-Rao
contour sets are circles. 

These contour shapes show that the Fisher-Rao distance
is not a perfect proxy for discriminability, since it
does not capture the interactions between the two
parameters $\log_{10}\sigma_1^2$ and $\log_{10}\sigma_2^2$ that lead to
the distinctive hexagonal shape of the accuracy and log-odds correct plots.
The Fisher-Rao distance provides a good approximation of the log-odds correct,
however, at smaller values of
$\log_{10}\sigma_1^2$ and $\log_{10}\sigma_2^2$.

To visualize this further, we plotted the values of the distances and
accuracies for three different slices of the 2D space, with fixed
values of $\log_{10}\sigma_2^2$ (0.0, 0.56, 1.12).
(Figure~\ref{fig:bayes2D}, middle row). For the first two slices
(black and red), the Fisher-Rao distance and the
log-odds correct are very similar across the range of values
of $\log_{10}\sigma_1^2$, but they become less similar
for the third slice (blue), where the asymmetry of
the log-odds correct is more pronounced.

In sum, the Fisher-Rao distance broadly captures the dependence of
discriminability on the distribution parameters in some scenarios,
particularly in the 1D case, and in the 2D case when the generalized eigenvalues
are closer to 1 (i.e.\ when $\log_{10}\sigma^2$ is close to 0),
but that it also fails to capture some patterns of
the dependence of discriminability on the parameters.

%%%%%%%%%%%%%%%%%%%%%%%%%%%%%%%%%%%%%%%%%%%%%%%%%%%%%%%%%%%%%%%%%%%%%%%%%%%
%%%%%%%%%%%%%%%%%%%%%%%%%%%%%%%%%%%%%%%%%%%%%%%%%%%%%%%%%%%%%%%%%%%%%%%%%%%
%%%%%%%%%%%%%%%%%%%%%%%%%%%%%%%%%%%%%%%%%%%%%%%%%%%%%%%%%%%%%%%%%%%%%%%%%%%

\section{Calvo-Oller bound as a surrogate for the Fisher-Rao distance}
\label{apd:calvo_oller}

In this section we discuss the Calvo-Oller bound in special cases
where we have exact formulas for the Fisher-Rao distance (see \cite{nielsen_simple_2023}).

\paragraph{Equal mean case.} For Gaussians with equal mean,
the Calvo-Oller bound is equal to the true Fisher-Rao distance.
First, we can use the affine-invariant property of the Calvo-Oller
bound \citep{nielsen_simple_2023} to subtract the common mean from
the data, thus reducing the problem to the zero-mean case.
Then, given that Calvo-Oller embedding for class $i$ is
$\Omegab_i = \begin{bmatrix} \Sigmab_i & \mathbf{0} \\
\mathbf{0}^T & 1 \\
\end{bmatrix}$,  it is easy to see that
$\Omegab_i^{-1}\Omegab_j = \begin{bmatrix} \Sigmab_i^{-1}\Sigmab_j & \mathbf{0} \\
\mathbf{0}^T & 1 \\ \end{bmatrix}$, and that the
generalized eigenvalues of $(\Omegab_i,\Omegab_j)$
are the eigenvalues of $\Sigmab_i^{-1}\Sigmab_j$ plus an
additional generalized eigenvalue equal to $1$. Since both
the Calvo-Oller distance and the Fisher-Rao distance are given by the sum of
the squared logarithm of the generalized eigenvalues (Equation~\ref{eq:ai_dist}),
it follows that the Calvo-Oller distance is equal to the true Fisher-Rao distance.

\begin{figure*}[t]
\begin{center}
\centerline{\includegraphics[width=0.3\textwidth]{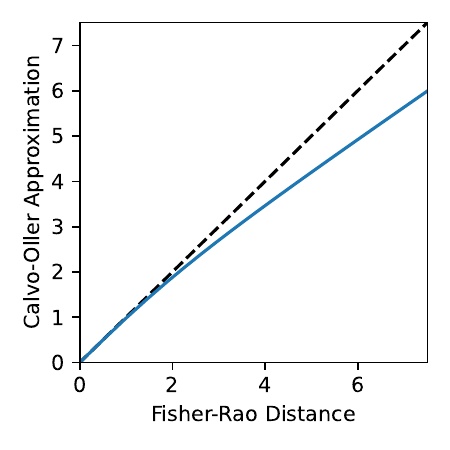}}
\caption{Calvo-Oller bound vs.\@ true Fisher-Rao distance
for the equal-covariance case. The Fisher-Rao distance and
the Calvo-Oller distance were computed for a range
of Mahalanobis distances ranging from 0 to 20.
The dashed line shows the identity line.
}
\label{fig:calvo_oller_same}
\end{center}
\end{figure*}

\paragraph{Equal covariance case.} For two Gaussians with equal
covariance, $\mathcal{N}(\mub_i,\Sigmab)$ and $\mathcal{N}(\mub_j,\Sigmab)$,
we denote the squared Mahalanobis distance as
\begin{equation}
  \label{eq:mahalanobis}
  d_{M}(\mub_i,\mub_j)^2 = (\mub_i - \mub_j)^T \Sigmab^{-1} (\mub_i - \mub_j)
\end{equation}
Then, the exact Fisher-Rao distance is given by \citep{nielsen_simple_2023}
\begin{equation}
  \label{eq:fr_equal_cov}
  d_{FR}(\mub_i,\mub_j) = \sqrt{2} \mathrm{arccosh}\left(1 + \frac{1}{4}d_{M}(\mub_i,\mub_j)^2\right)
\end{equation}
Figure~\ref{fig:calvo_oller_same} shows the Calvo-Oller bound
as a function of the true Fisher-Rao distance for Gaussians
with identity covariance, ranging from Mahalanobis distance of 0 to 20.
Across this large range of Mahalanobis distances, the Calvo-Oller bound is close
to the true Fisher-Rao distance.

\section{LDA features maximize squared Mahalanobis distances}
\label{apd:lda}

Here, we prove that Linear Discriminant Analysis (LDA)
maximizes the squared Mahalanobis distances between classes
when the classes are homoscedastic Gaussians. 
The Mahalanobis distance equals the Fisher-Rao distance
along the submanifold of Gaussians with equal
covariance\footnote{This is not the same as the Fisher-Rao distance
along the general manifold of Gaussian distributions.}, which
means that LDA maximizes the pairwise Fisher-Rao
(squared) distances between classes along that submanifold.

For a labeled random variable $\mathbf{x} \in \mathbb{R}^n$ with
class labels $y \in \{1,\ldots,c\}$, the goal of LDA is
to find the filters $\F \in \mathbb{R}^{n \times m}$ such that
the variable $\mathbf{z} = \F^T\mathbf{x}$ maximizes 
the between-class scatter relative to the within-class scatter.
This is typically formulated as maximizing the
Fisher criterion
\begin{equation}
  \label{eq:Fisher}
  \mathrm{Tr}\left( \Sigmab^{-1}\mathbf{S}_{\mathbf{z}} \right)
\end{equation}
where $\Sigmab$ is the residual within-class covariance
matrix of the data (i.e.\ the covariance matrix
of the data after subtracting the class mean from each data point),
and $\mathbf{S}_{\mathbf{z}}$ is the between-class scatter matrix
of $\mathbf{z}$, defined as
\begin{equation*}
  \mathbf{S}_{\mathbf{z}} = \frac{1}{c} \sum_{i=1}^c (\mub_i - \mub)(\mub_i - \mub)^T
\end{equation*}
where $\mub_i$ is the mean of class $i$ in the feature space and
$\mub = \sum_{i=1}^c\mub_i$.

The sum of pairwise squared Mahalanobis distances between the classes is given by
\begin{equation}
  \label{eq:LDAdist}
  \frac{1}{2}\sum_{i=1}^c\sum_{j=1}^c d_{M}(\mub_i, \mub_j)^2 =
  \frac{1}{2}\sum_{i=1}^c\sum_{j=1}^c (\mub_i - \mub_j)^T \Sigmab^{-1} (\mub_i - \mub_j)
\end{equation}
where the factor of $1/2$ is included to control for double-counting.

We can show that the objectives in Equation~\ref{eq:Fisher}
and Equation~\ref{eq:LDAdist} are equivalent. First, we assume without loss
of generality that the overall mean $\mub = \sum_{i=1}^c \mub_i = 0$. Then, we have
\begin{equation*}
  \begin{split}
  \frac{1}{2}\sum_{i=1}^c\sum_{j=1}^c d_{M}(\mub_i, \mub_j)^2 
  &= \frac{1}{2}\sum_{i=1}^c\sum_{j=1}^c (\mub_i - \mub_j)^T \Sigmab^{-1} (\mub_i - \mub_j) \\
  &= \frac{1}{2}\sum_{i=1}^c\sum_{j=1}^c \left( \mub_i^T \Sigmab^{-1} \mub_i + \mub_j^T \Sigmab^{-1} \mub_j -
    2\mub_i^T \Sigmab^{-1} \mub_j \right) \\
  &= \sum_{i=1}^c \mub_i^T \Sigmab^{-1} \mub_i -
    \sum_{i=1}^c\sum_{j=1}^c \mub_i^T \Sigmab^{-1} \mub_j \\
    &= \sum_{i=1}^c \mub_i^T \Sigmab^{-1} \mub_i -
    \left(\sum_{i=1}^c \mub_i^T \right) \Sigmab^{-1} \left( \sum_{j=1}^c \mub_j\right) \\
  &= \sum_{i=1}^c \mub_i^T \Sigmab^{-1} \mub_i \\
  &= \mathrm{Tr}\left( \Sigmab^{-1} \sum_{i=1}^c \mub_i \mub_i^T \right) \\
  &= c\mathrm{Tr}\left( \Sigmab^{-1} \mathbf{S}_\mathbf{z} \right) \\
  \end{split}
\end{equation*}
In the first five lines we expanded the squared Mahalanobis distance,
used the linearity of the dot product and the fact that $\mub = 0$.
In the last two lines we used the linearity and the cyclic property
of the trace, and that $\sum_{i=1}^c \mub_i \mub_i^T = c\mathbf{S}_\mathbf{z}$
because $\mub = 0$. This shows that maximizing the LDA criterion
maximizes the pairwise squared Mahalanobis distances between classes.

%%%%%%%%%%%%%%%%%%%%%%%%%%%%%%%%%%%%%%%%%%%%%%%%%%%%%%%%%%%%%%%%%%%%%%%%%%%
%%%%%%%%%%%%%%%%%%%%%%%%%%%%%%%%%%%%%%%%%%%%%%%%%%%%%%%%%%%%%%%%%%%%%%%%%%%
%%%%%%%%%%%%%%%%%%%%%%%%%%%%%%%%%%%%%%%%%%%%%%%%%%%%%%%%%%%%%%%%%%%%%%%%%%%

\section{Comparison of different distances}
\label{apd:distances}

In this section, we compare the behavior of the Fisher-Rao distance with
the two other main dissimilarity measures in this work: the Hellinger and
the Bhattacharyya distance\footnote{We remind the reader that the
Bhattacharyya distance is not actually a distance in the geometric sense.
We still refer to it as a distance, in line with the literature.}.

\subsection{Distances scale differently in the equal-covariance case}
\label{apd:distances_equal_cov}

For the case of Gaussians with equal covariance and different means,
all three distances have closed form expressions as a function of the
Mahalanobis distance (Equation~\ref{eq:fr_equal_cov} and Equation~\ref{eq:bhattacharyya}).

Figure~\ref{fig:distances_comp} in the main text shows that the three distances
grow monotonically with the Mahalanobis distance, but with different
growth rates. The Bhattacharyya distance equals the squared Mahalanobis
distance in the equal covariance case (see Equation~\ref{eq:bhattacharyya}).
As two classes become more separated, the Bhattacharyya distance grows faster.
On the other hand, the Fisher-Rao distance grows sublinearly with the Mahalanobis
distance. As two classes become more separated, the Fisher-Rao distance
grows more slowly. Finally, the Hellinger distance is bounded 
unlike the two other distances. As two classes become more separated,
the Hellinger distance approaches the maximum value of 1.

The difference in the growth rates has implications
for their behavior as learning objectives.
In the multi-class dimensionality reduction problem,
the Bhattacharyya distance will favor separating classes
that are already well-separated, since those lead to a larger
increase in the average pairwise Bhattacharyya distance.
On the other hand, the Hellinger distance will strongly favor
separating classes that are close together, since further
separating classes that are already far apart will barely
change the average Hellinger distance.
The Fisher-Rao distance will behave somewhere in between:
it favors separating classes that are close together,
but it is still sensitive to classes that are farther apart, since it
is an unbounded distance.

We designed a toy dataset to illustrate how the different
relative weights of the pairs of classes for the different
objectives can affect the learning outcome. 
We learned one filter with SQFA and with LDA for this dataset.
Because LDA maximizes the average squared Mahalanobis distance
(Appendix~\ref{apd:lda}), it is equivalent
to maximizing the Bhattacharyya distance in the equal-covariance case.

\begin{figure*}[t]
\begin{center}
\centerline{\includegraphics[width=\textwidth]{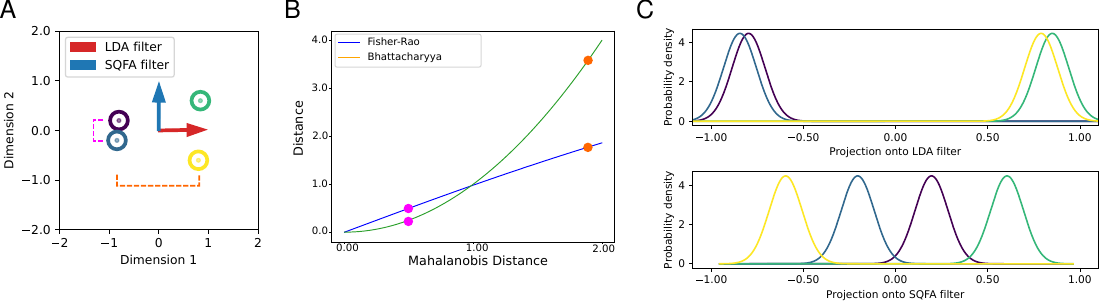}}
\caption{\textbf{A)} Toy dataset with 4 classes of 2D Gaussian distributions with
identical covariance and different means. Arrows indicate the filters
learned by LDA (red) and by SQFA (blue).
\textbf{B)} Value of the Fisher-Rao distance (blue curve) and
Bhattacharyya distance (green curve) for
the distance between clusters along the horizontal axis (orange dot)
and the spacing along classes along the vertical axis (magenta dot).
\textbf{C} Projection of the classes onto the LDA (top) and SQFA (bottom) filters.
}
\label{fig:sqfa_vs_lda}
\end{center}
\end{figure*}

The dataset consists of 4 classes of 2D Gaussian distributions with
the same covariance and different means (Figure~\ref{fig:sqfa_vs_lda}A).
Along the horizontal axis, the classes form two clusters that are
far from each other, but that have high within-cluster overlap.
Along the vertical axis, the classes are evenly separated at a short
distance from each other, but with little overlap.
The large distances along the horizontal axis will be highly
weighted by the Bhattacharyya distance (i.e. LDA), whereas they
will have a more modest contribution to the Fisher-Rao distance
(Figure~\ref{fig:sqfa_vs_lda}B). The filter learned by LDA
aligns with the horizontal axis, whereas the first filter learned by SQFA
aligns with the vertical axis (Figure~\ref{fig:sqfa_vs_lda}A). 
The classes have less overlap along the vertical axis
(Figure~\ref{fig:sqfa_vs_lda}C), indicating that in this case,
SQFA leads to better class separation than LDA.
This example illustrates how the different distances can lead to
different learning outcomes.

%%%%%%%%%%%%%%%%%%%%%%%%%%%%%%%%%%%%%%%%%%%%%%%%%%%%%%%%%%%%%%%%%%%%%%%%%%%

\subsection{Distances scale differently in the equal-mean case}
\label{apd:distances_equal_mean}

Next, we consider the case of two Gaussians with zero mean
and different covariances. In Figure~\ref{fig:distances_comp} of the
main text we showed how the distances scale for the 1D case. Here we
complement that analysis by showing how the distances scale for the 2D case.
We examine 2D distributions with diagonal covariance matrices,
that is, $\Sigmab_i = \mathbf{D} = \mathrm{diag}(\sigma_1^2,\sigma_2^2)$
and $\Sigmab_j = \mathbf{I}$. Because of the affine invariance of all
three distances, the general problem can be reduced to this special case.

\begin{figure*}[t]
\begin{center}
\centerline{\includegraphics[width=0.8\textwidth]{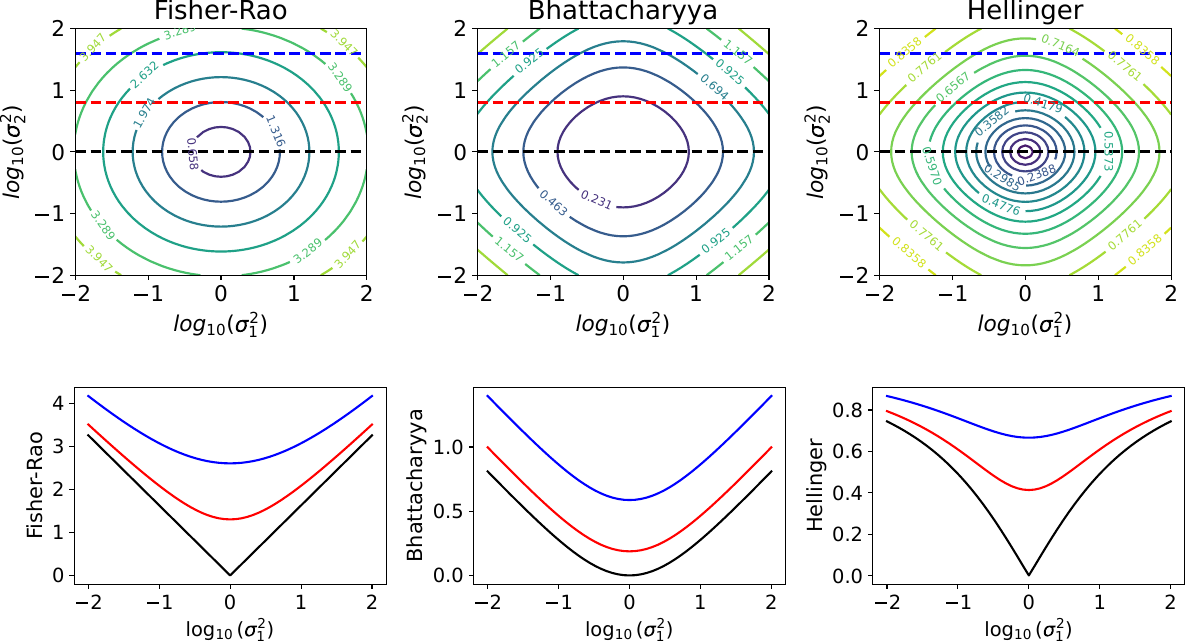}}
\caption{Distances as a function of covariance differences.
  The three distances (Fisher-Rao, Hellinger, Bhattacharyya)
  are shown as a function of $\log_{10}(\sigma_1^2)$ and
  $\log_{10}(\sigma_2^2)$, where the two Gaussian distributions
  are $\mathcal{N}(0,\mathbf{D})$ and $\mathcal{N}(0,\mathbf{I})$.
  The top row shows contour plots of the distances,
  while the bottom row shows 1D slices of the distances.
  The values used for the slices are shown as dashed lines
  in the top row.
}
\label{fig:distances_comp2}
\end{center}
\end{figure*}

Again, the three distances behave differently (Figure~\ref{fig:distances_comp2}).
The Hellinger distance is bounded between 0 and 1, while
the Bhattacharyya and Fisher-Rao distances are unbounded.
One interesting pattern is that whereas the contours of
the Fisher-Rao distance are circles (as expected from
its formula), the contours of the Hellinger and Bhattacharyya distances
are more circular closer to the origin, but acquire a rhomboid 
shape farther from the origin.

The bottom plots show the three distances as a function of
$\log_{10}(\sigma_1^2)$ for different fixed values of
$\log_{10}(\sigma_2^2)$. This plot illustrates how the
effect of one generalized eigenvalue depends on the value of the
other generalized eigenvalue.
We see that while the effect of $\log_{10}(\sigma_1^2)$ 
depends on the value of $\log_{10}(\sigma_2^2)$ for the Fisher-Rao
and Hellinger distances, the two
are independent for the Bhattacharyya distance. 
It is interesting to note that, of the different distances, the slices
of the Fisher-Rao distance follow most closely the slices of the
log-odds correct shown in Figure~\ref{fig:bayes2D}.

In sum, these examples illustrate some of the differences
between the Fisher-Rao distance and other distances, in
the special cases with closed-form expressions.
The different behaviors of these distances mean that they will
put different weights on the different pairwise distances
between classes, leading to potentially different learning outcomes.

%%%%%%%%%%%%%%%%%%%%%%%%%%%%%%%%%%%%%%%%%%%%%%%%%%%%%%%%%%%%%%%%%%%%%%%%%%%
%%%%%%%%%%%%%%%%%%%%%%%%%%%%%%%%%%%%%%%%%%%%%%%%%%%%%%%%%%%%%%%%%%%%%%%%%%%
%%%%%%%%%%%%%%%%%%%%%%%%%%%%%%%%%%%%%%%%%%%%%%%%%%%%%%%%%%%%%%%%%%%%%%%%%%%

\section{Variability of results across runs}
\label{apd:variability}

In this section we show the variability of the SQFA results across training runs.

In the experiments of the main text, except for the neural dataset, we ran each SQFA
variant 10 times with different random seeds, and reported the median performance across runs.
Here we report the variability across runs as the central 80\% quantile range,
that is, the difference between the 90\% and the 10\% quantiles across runs.
To include the variability of the neural dataset, we used a new set of
training runs, where we kept the neural recording session and train-test split
fixed, and ran each SQFA variant 10 times on this single dataset. This way,
the results below reflect SQFA training variability, not dataset variability.

We see that for most methods and datasets, the variability across runs is 
extremely low, of the order of 0.1\% or less. For some conditions
there is larger variability, especially when two filters are learned.
For most cases, this variability comes from a few runs that found a poor local
minimum, and so it does not change the overall pattern of results
reported in the main text. The practical implication of this variability
is that for most conditions random initialization does not seem to have
a relevant effect on the final performance, but that for some conditions,
especially when learning only two filters, it may be advisable to run the method
multiple times with different random seeds and select the best-performing run.

% table_csvs/svhn_n_filters_qda_review_wide_q10_q90_width.csv
\begin{table}[h]
  \centering
  \small
  \setlength{\tabcolsep}{4pt}
  \caption{QDA accuracy 10--90\% quantile range width over runs (as \% correct).}
  \label{tab:svhn_n_filters_qda_q10_q90}
  \begin{tabular}{rccccccc}
    \toprule
    $m$ & Dataset & SQFA & smSQFA & SQFA-H & SQFA-B & SQFA-W & SQFA-J \\
    \midrule
    2 & SVHN & 3.5 & 4.4 & 6.7 & 3.9 & 0.1 & 0.0 \\
    4 & SVHN &  0.0 & 0.0 & 0.1 & 2.2 & 0.0 & 0.0 \\
    8 & SVHN &  2.5 & 0.1 & 0.0 & 4.0 & 0.0 & 0.0 \\
    16 & SVHN &  0.0 & 0.0 & 0.0 & 0.0 & 0.0 & 0.0 \\
    2 & MNIST & 8.2 & 4.1 & 2.6 & 0.2 & 0.6 & 0.0 \\
    4 & MNIST & 1.0 & 0.0 & 0.1 & 0.0 & 0.0 & 0.0 \\
    8 & MNIST & 0.0 & 0.1 & 0.1 & 0.4 & 0.0 & 0.2 \\
    16 & MNIST & 0.0 & 0.1 & 0.0 & 0.1 & 0.0 & 0.4 \\
    2 & Speed & 0.6 & 0.1 & 0.0 & 50.2 & 0.0 & 0.0 \\
    4 & Speed & 0.0 & 0.0 & 0.1 & 0.1 & 0.8 & 0.0 \\
    6 & Speed & 1.6 & 1.4 & 0.5 & 0.0 & 3.1 & 0.0 \\
    8 & Speed & 0.0 & 0.0 & 0.1 & 0.3 & 0.0 & 0.0 \\
    2 & Neural data& 0.0 & 0.0 & 0.0 & 0.0 & 0.0 & 0.0 \\
    4 & Neural data & 0.0 & 0.0 & 0.3 & 0.0 & 0.0 & 0.0 \\
    8 & Neural data & 2.2 & 0.0 & 0.6 & 0.0 & 0.0 & 0.0 \\
    \bottomrule
  \end{tabular}
\end{table}

%%%%%%%%%%%%%%%%%%%%%%%%%%%%%%%%%%%%%%%%%%%%%%%%%%%%%%%%%%%%%%%%%%%%%%%%%%%
%%%%%%%%%%%%%%%%%%%%%%%%%%%%%%%%%%%%%%%%%%%%%%%%%%%%%%%%%%%%%%%%%%%%%%%%%%%
%%%%%%%%%%%%%%%%%%%%%%%%%%%%%%%%%%%%%%%%%%%%%%%%%%%%%%%%%%%%%%%%%%%%%%%%%%%

\section{Robustness to the regularization parameter}
\label{apd:regularization_sensitivity}

In the main text, we selected the regularization parameter for the different
SQFA variants using a validation set (except for the speed estimation
dataset, where we used a parameter derived from the literature \cite{burge_optimal_2015}).
Here we test the robustness of the method to the choice of the regularization parameter.

\begin{figure*}[h]
\begin{center}
\centerline{\includegraphics[width=1.0\textwidth]{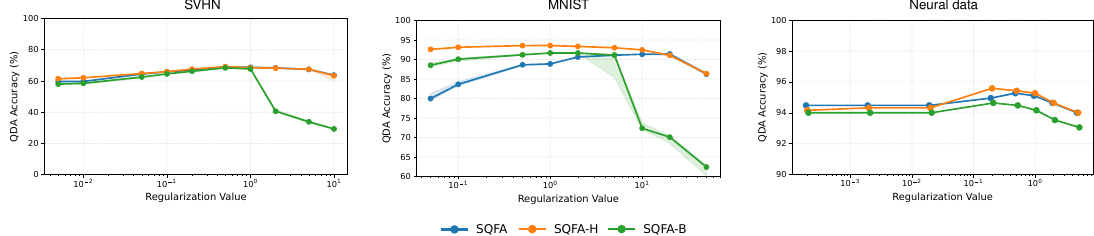}}
\caption{
Classification accuracy as a function of the regularization parameter.
The accuracy of a QDA classifier trained on the features learned by SQFA,
SQFA-H and SQFA-B is shown as a function of the regularization parameter $\lambda$,
for the SVHN (left), MNIST (center), and neural recordings (right) datasets.
The values of $\lambda$ are shown in log scale.}
\label{fig:regularization}
\end{center}
\end{figure*}

We learned 9 filters with SQFA, SQFA-H and SQFA-B using a range of values of
the regularization parameter $\lambda$, for the SVHN and MNIST datasets,
and 2 filters for the neural recordings dataset (we used only one of the
neural recordings session, since different sessions have different
optimal $\lambda$ values).
All three methods showed a considerable degree of robustness to the
choice of $\lambda$ (Figure~\ref{fig:regularization}). 
SQFA-H showed the highest robustness, with similar performance over a range
of $\lambda$ values spanning an order of magnitude or more. SQFA and SQFA-B
showed less robustness than SQFA-H, but still showed good
performance over a large range of $\lambda$ values for all datasets.
These results show that the performance, and the results in the main
paper are not highly sensitive to the choice of the regularization parameter.
For practical applications, it means that while selecting the regularization parameter
using a validation set is advisable, this parameter does not require tight fine tuning.

%%%%%%%%%%%%%%%%%%%%%%%%%%%%%%%%%%%%%%%%%%%%%%%%%%%%%%%%%%%%%%%%%%%%%%%%%%%
%%%%%%%%%%%%%%%%%%%%%%%%%%%%%%%%%%%%%%%%%%%%%%%%%%%%%%%%%%%%%%%%%%%%%%%%%%%
%%%%%%%%%%%%%%%%%%%%%%%%%%%%%%%%%%%%%%%%%%%%%%%%%%%%%%%%%%%%%%%%%%%%%%%%%%%

\section{Complexity analysis and training times}

\subsection{Computational complexity of SQFA}
\label{apd:complexity}

Here we analyze the computational complexity of SQFA.
Specifically, we analyze the cost of taking a gradient step
using the SQFA objective.
For simplicity, we analyze the case of smSQFA
(assuming zero means), but this
is equivalent to the case of SQFA, since the operations
are the same except substituting the $m \times m$ matrices
with $(m+1) \times (m+1)$ matrices.

The first step is to compute the class covariances
in the feature space. As mentioned in the Methods section,
we achieve this by transforming the covariances of the raw data using the formula
$\Sigmab_i = \F^T\Phib_i\F$,
where $\Phib_i$ is the covariance of the data for class $i$.
The matrix $\Phib_i$ is $n \times n$, and the matrix $\F$ is
$n \times m$, where $n$ is the dimensionality of the data
and $m$ the number of filters. The cost of computing $\Sigmab_i$
for all $c$ classes is $O(cmn^2)$.

Next, we compute the generalized eigenvalues of the
pair $(\Sigmab_i,\Sigmab_j)$, where the matrices are
$m \times m$. The cost of computing 
$\Sigmab_i^{-1}$ is $O(m^3)$.
Then, we compute the product matrix $\Sigmab_i^{-1}\Sigmab_j$,
which also costs $O(m^3)$.
Finally, the cost of computing the eigenvalues of $\Sigmab_i^{-1}\Sigmab_j$
is also $O(m^3)$. Performing this for all pairs of
classes has a complexity of $O(c^2m^3)$.

The cost of computing the gradient for the three operations described above
is the same as the cost of computing the operations themselves.
Thus, the complexity of taking a gradient step in SQFA is $O(c^2m^3 + cmn^2)$.

It is important to note that the most expensive operations
occur in the feature space, and 
the dimensionality of the feature space, $m$, is typically much smaller
than the dimensionality of the data, $n$.

%%%%%%%%%%%%%%%%%%%%%%%%%%%%%%%%%%%%%%%%%%%%%%%%%%%%%%%%%%%%%%%%%%%%%%%%%%%

\subsection{Scaling analysis}
\label{apd:scaling}

Finally, we analyzed how SQFA scales with the data dimensionality
and with the number of learned filters. We used five-class synthetic
datasets in which all discriminative information was contained in a
known informative subspace.  The informative subspace was built by
combining a subspace of dimension 4 with differences in the class
means, and a complementary subspace where the classes had the same
means but covariances with a minor and a major axis rotated with
respect to one another. Thus, the informative subspace combined
first- and second-order differences between classes.
The rest of the dimensions had zero mean and identity covariance.
The resulting space was rotated so that the informative subspace was
not aligned with the canonical axes.
We compared QDA accuracy using SQFA features
against QDA accuracy using the true informative subspace.

For the first analysis, we generated different datasets, but for
each we fixed the informative subspace to eight dimensions and
embedded it in ambient spaces of increasing dimensionality. SQFA
learned eight filters in each dataset. Across ambient dimensions up to
$20{,}000$, SQFA matched the accuracy obtained from the true informative
subspace, with fit times ranging from seconds to minutes on a consumer
laptop (Table~\ref{tab:scaling_dimensions}).

\begin{table}[ht]
  \centering
  \caption{Scaling with ambient dimensionality. SQFA learned eight filters
  in a five-class synthetic problem with an eight-dimensional informative
  subspace.}
  \begin{tabular}{rrrr}
    \toprule
    Ambient dimension & Max accuracy (\%) & SQFA accuracy (\%) & Fit time (s) \\
    \midrule
    1000  & 92.5 & 92.5 & 0.8 \\
    2000  & 92.5 & 92.6 & 2.6 \\
    5000  & 92.4 & 92.4 & 7.8 \\
    10000 & 92.7 & 92.6 & 30.7 \\
    20000 & 92.6 & 92.5 & 176.2 \\
    \bottomrule
  \end{tabular}
  \label{tab:scaling_dimensions}
\end{table}

Second, we fixed the ambient dimensionality at $5000$ and increased
the dimension of the informative subspace. For each dataset, SQFA
learned as many filters as there were informative dimensions. SQFA
features achieved accuracy close to that obtained from the true
informative subspace, including when learning hundreds of filters
(Table~\ref{tab:scaling_filters}).

\begin{table}[ht]
  \centering
  \caption{Scaling with the number of learned filters. The ambient
  dimensionality was fixed at $5000$, and the number of filters matched
  the dimension of the informative subspace.}
  \begin{tabular}{rrrr}
    \toprule
    $N$ filters & Max accuracy (\%) & SQFA accuracy (\%) & Fit time (s) \\
    \midrule
    10  & 65.9 & 65.9 & 10.8 \\
    50  & 76.0 & 73.2 & 13.8 \\
    100 & 77.5 & 74.6 & 22.2 \\
    500 & 64.9 & 63.7 & 85.0 \\
    \bottomrule
  \end{tabular}
  \label{tab:scaling_filters}
\end{table}

These examples show that SQFA can scale to tens of thousands of
ambient dimensions and hundreds of filters in controlled synthetic
settings. We note that here we used the known synthetic class statistics,
but in real datasets, performance will also depend on how
accurately the class statistics can be estimated. In the implementation
used here, memory is eventually limited by storing dense class
covariance matrices $\Phib_i$; sparse covariance representations,
optimization from raw data, or a preliminary PCA step could extend
the practical scaling.

\subsection{Training times}
\label{apd:training_times}

For each dataset, we recorded the time it took to train the models
on a consumer laptop with a 12th Gen Intel Core i7-1270P
with 32 GB of RAM.

The training times of SQFA and its variants were in the order
of seconds for all datasets (Figure~\ref{tab:training_time}).
Remarkably, the training time for SQFA was comparable to that
of the fastest training methods\@.

\begin{table}[t]
  \centering
  \small
  \setlength{\tabcolsep}{4pt}
  \caption{Training time (s).}
  \label{tab:training_time}
  \begin{tabular}{lrcccccccccccc}
    \toprule
    $m$ & Dataset & SQFA & smSQFA & SQFA-H & SQFA-B & SQFA-W & SQFA-J & LDA & SPCA & LFDA & WDA & LMNN & PCA \\
    \midrule
    8 & SVHN & 7.5 & 9.2 & 10.1 & 92.9 & 7.5 & 17.1 & 3.2 & 8.9 & 13.2 & 10.4 & 2340.0 & 1.9 \\
    8 & MNIST & 1.2 & 6.8 & 1.3 & 3.5 & 4.0 & 7.8 & 1.9 & 4.2 & 6.4 & 68.3 & 385.1 & 1.1 \\
    8 & Speed & 3.0 & 3.9 & 1.4 & 11.7 & 5.4 & 6.0 & 0.6 & 0.5 & 0.6 & 275.6 & 1936.8 & 1.2 \\
    8 & Neural data & 0.6 & 0.6 & 0.4 & 0.5 & 0.3 & 0.7 & -- & 0.1 & 0.0 & 34.6 & 10.7 & 0.0 \\
    \bottomrule
  \end{tabular}
\end{table}
%%%%%%%%%%%%%%%%%%%%%%%%%%%%%%%%%%%%%%%%%%%%%%%%%%%%%%%%%%%%%%%%%%%%%%%%%%%
%%%%%%%%%%%%%%%%%%%%%%%%%%%%%%%%%%%%%%%%%%%%%%%%%%%%%%%%%%%%%%%%%%%%%%%%%%%
%%%%%%%%%%%%%%%%%%%%%%%%%%%%%%%%%%%%%%%%%%%%%%%%%%%%%%%%%%%%%%%%%%%%%%%%%%%

\section{Evaluation with kNN classifiers}
\label{apd:knn}

In the main text, we evaluated the performance of the different methods
by training a quadratic discriminant analysis (QDA) classifier on the
learned features. Here, we show that the same patterns of results are obtained
when using a k-nearest neighbor (kNN) classifier instead of QDA
(Tables~\ref{tab:svhn_n_filters_knn}-\ref{tab:zand_kohn_n_filters_knn}).
All training and evaluation procedures were the same as in the main text, except that
instead of training a QDA classifier, we trained a kNN classifier with $k=5$.

Broadly, the results are similar to those obtained with QDA.
For SVHN and MNIST, SQFA-H remained in general the best-performing method,
and SQFA and SQFA-B also performed very well, among the top performing
methods. It is remarkable that SQFA and SQFA-H remained highly competitive
when using a kNN classifier, even when compared to methods that are considered
to be better suited for kNN, such as LFDA, WDA and LMNN.

For the speed estimation dataset, there is some interesting effect of
dimensionality, where the performance of kNN drops as the number of
filters increases for most SQFA variants. The one exception is
SQFA-J, but its special behavior can be explained by looking at its
filters in Figure~\ref{fig:speed}. The figure shows that
SQFA-J learns coarsely the same pair of filters multiple times.
For this dataset, this is known to happen when the learning procedure
prioritizes minimizing the effects of internal noise (i.e. the regularization term) on
a fixed pair of features rather than extracting new features \cite{burge_accuracy_2017}.
Thus, by learning redundant features, the feature space of SQFA-J remains
low-dimensional even as we increase the number of filters, avoiding the negative effects of
increasing the dimensionality. Conversely, the methods that are not
SQFA variants have a low performance for 2 filters, but their performance
increases with the number of filters, which is the expected behavior for
these methods that are more suited for kNN\@.

Overall, these results show that the patterns of results obtained with QDA are
not specific to that classifier, and that they generalize to a different
type of classifier. This suggests that the SQFA variants should be a useful
class of methods even when classifiers other than QDA are used.

% table_csvs/svhn_n_filters_knn_review_wide_median.csv
\begin{table}[h]
  \centering
  \small
  \setlength{\tabcolsep}{4pt}
  \caption{SVHN KNN classification accuracy (median percentage over runs).}
  \label{tab:svhn_n_filters_knn}
  \begin{tabular}{rcccccccccccc}
    \toprule
    $m$ & SQFA & smSQFA & SQFA-H & SQFA-B & SQFA-W & SQFA-J & LDA & SPCA & LFDA & WDA & LMNN & PCA \\
    \midrule
    2 & \textbf{33.9} & 32.1 & \textbf{33.5} & 31.6 & 26.5 & 30.8 & 16.7 & 13.8 & 17.9 & 17.4 & 14.3 & 14.3 \\
    4 & \textbf{54.3} & 54.2 & \textbf{54.4} & 53.7 & 45.8 & 53.5 & 21.3 & 19.1 & 25.4 & 35.2 & 20.7 & 19.8 \\
    8 & \textbf{69.3} & \textbf{69.2} & \textbf{69.3} & 66.0 & 52.1 & 68.6 & 31.3 & 31.0 & 33.3 & 51.1 & 43.4 & 29.0 \\
    16 & \textbf{78.0} & 75.7 & \textbf{78.1} & 77.6 & 57.5 & 76.8 & -- & 37.0 & 41.1 & 64.9 & 57.7 & 44.6 \\
    \bottomrule
  \end{tabular}
\end{table}

% table_csvs/mnist_n_filters_knn_review_wide_median.csv
\begin{table}[h]
  \centering
  \small
  \setlength{\tabcolsep}{4pt}
  \caption{MNIST KNN classification accuracy (median percentage over runs).}
  \label{tab:mnist_n_filters_knn}
  \begin{tabular}{rcccccccccccc}
    \toprule
    $m$ & SQFA & smSQFA & SQFA-H & SQFA-B & SQFA-W & SQFA-J & LDA & SPCA & LFDA & WDA & LMNN & PCA \\
    \midrule
    2 & 57.2 & \textbf{57.7} & \textbf{62.9} & 51.9 & 48.4 & 45.9 & 52.3 & 44.2 & 54.9 & 50.5 & 47.2 & 42.4 \\
    4 & 79.9 & 77.1 & \textbf{86.7} & \textbf{82.6} & 74.3 & 76.1 & 82.0 & 71.3 & 81.9 & 67.2 & 73.2 & 63.3 \\
    8 & 91.8 & 90.8 & \textbf{93.9} & 92.1 & 91.4 & 89.9 & 92.0 & 90.9 & 90.8 & 90.1 & \textbf{92.7} & 90.1 \\
    16 & 96.0 & 96.2 & \textbf{96.6} & 96.1 & \textbf{96.3} & 94.2 & -- & 94.9 & 93.7 & 96.1 & \textbf{96.3} & \textbf{96.3} \\
    \bottomrule
  \end{tabular}
\end{table}

% table_csvs/motion_n_filters_knn_review_wide_median.csv
\begin{table}[h]
  \centering
  \small
  \setlength{\tabcolsep}{4pt}
  \caption{Speed KNN classification accuracy (median percentage over runs).}
  \label{tab:motion_n_filters_knn}
  \begin{tabular}{rccccccccccccc}
    \toprule
    $m$ & SQFA & smSQFA & SQFA-H & SQFA-B & SQFA-W & SQFA-J & AMA & LDA & SPCA & LFDA & WDA & LMNN & PCA \\
    \midrule
    2 & 53.0 & 53.3 & 51.3 & 53.1 & 30.7 & \textbf{53.9} & \textbf{56.3} & 4.3 & 7.0 & 4.7 & 6.2 & 3.4 & 23.3 \\
    4 & 29.8 & 29.5 & 30.4 & 28.4 & 23.3 & \textbf{59.7} & \textbf{36.0} & 8.9 & 18.1 & 17.1 & 14.3 & 10.6 & 19.8 \\
    6 & 22.0 & 23.4 & 25.6 & 20.8 & 21.8 & \textbf{61.7} & \textbf{29.1} & 15.7 & 17.2 & 8.0 & 28.3 & 17.0 & 16.2 \\
    8 & 23.5 & 23.4 & 26.6 & 18.7 & 20.4 & \textbf{63.3} & \textbf{29.6} & 19.5 & 17.1 & 25.2 & 29.0 & 22.0 & 17.4 \\
    \bottomrule
  \end{tabular}
\end{table}

% table_csvs/zand_kohn_n_filters_knn_review_wide_mean.csv
\begin{table}[h]
  \centering
  \small
  \setlength{\tabcolsep}{4pt}
  \caption{Zand-Kohn KNN classification accuracy (mean percentage over runs).}
  \label{tab:zand_kohn_n_filters_knn}
  \begin{tabular}{rcccccccccccc}
    \toprule
    $m$ & SQFA & smSQFA & SQFA-H & SQFA-B & SQFA-W & SQFA-J & LDA & SPCA & LFDA & WDA & LMNN & PCA \\
    \midrule
    2 & 89.4 & 86.9 & \textbf{90.3} & 86.9 & 84.6 & 85.6 & 87.2 & 84.6 & 86.9 & 86.2 & \textbf{90.2} & 65.3 \\
    4 & 93.7 & 94.4 & \textbf{95.6} & 92.4 & 84.7 & 91.4 & 93.0 & 91.9 & 93.1 & 89.1 & \textbf{95.0} & 90.6 \\
    6 & 95.8 & 96.0 & \textbf{96.4} & 95.9 & 84.6 & 95.1 & -- & 96.1 & 96.4 & 84.3 & \textbf{96.2} & 95.6 \\
    \bottomrule
  \end{tabular}
\end{table}

%%%%%%%%%%%%%%%%%%%%%%%%%%%%%%%%%%%%%%%%%%%%%%%%%%%%%%%%%%%%%%%%%%%%%%%%%%%
%%%%%%%%%%%%%%%%%%%%%%%%%%%%%%%%%%%%%%%%%%%%%%%%%%%%%%%%%%%%%%%%%%%%%%%%%%%
%%%%%%%%%%%%%%%%%%%%%%%%%%%%%%%%%%%%%%%%%%%%%%%%%%%%%%%%%%%%%%%%%%%%%%%%%%%

\section{Regularization and invariance}
\label{apd:regularization}

\subsection{Invariance to invertible linear transformations}

To better understand SQFA, it is important to consider the
invariance properties of the Fisher-Rao and the affine-invariant
distances.

First, we consider the affine-invariant distance and the
case of smSQFA\@.
If $\mathbf{G} \in GL(m)$ where $GL(m)$ is the
General Linear group, composed of non-singular $m$-by-$m$ matrices, then
\begin{equation}
  \label{eq:invariance}
  d_{AI}(\Psib_i,\Psib_j) = d_{AI}(\mathbf{G}^T\Psib_i\mathbf{G}, \mathbf{G}^T\Psib_j\mathbf{G})
\end{equation}
In words, the affine-invariant distance is invariant to the action
by congruence of $GL(m)$. 

For a variable $\mathbf{z} \in \mathbb{R}^m$ with
second-moment matrix $\Psib_i$, the transformed matrix
$\Psib_i' = \mathbf{G}^T\Psib_i\mathbf{G}$ corresponds
to the second moment matrix of the transformed variable
$\mathbf{z}' = \mathbf{G}^T\mathbf{z}$.
Thus, the distance is invariant to invertible linear transformations
of the underlying variable $\mathbf{z}$.
Importantly, for the case of SQFA, where the variable $\mathbf{z}$
is obtained as $\mathbf{z} = \F^T\mathbf{x}$, this
is also equivalent to a transformation of the filters.
Specifically, if $\mathbf{z} = \F^T\mathbf{x}$,
then $\mathbf{z}' = \F'^T\mathbf{x}$, where
$\F' = \F\mathbf{G}$.

In the context of smSQFA, and in the absence of regularization
(see Methods section), if a set of filters are transformed
as $\F' = \F\mathbf{G}$, then the second-moment matrices
for all the classes will be transformed as
$\Psib_i' = \mathbf{G}^T\Psib_i\mathbf{G}$. Therefore,
according to Equation~\ref{eq:invariance}, the pairwise
distances between classes will be the same for both sets
of filters. In other words, the objective function of smSQFA
is invariant to invertible linear transformations of the filters.

The Fisher-Rao distance and the Calvo-Oller bound
are also invariant to affine transformations of the data space
\cite{nielsen_elementary_2020}. This makes the filters learned
by SQFA (like smSQFA above) non-unique up to invertible linear
transformations (again, in the absence of regularization).
To see this, let
$\mathbf{H} = \begin{bmatrix}
  \mathbf{G} & \mathbf{0} \\
  \mathbf{0} & 1
\end{bmatrix}$, where $\mathbf{0}$ is a $m$-by-1 vector of zeros,
and $\mathbf{H} \in GL(m+1)$.
We denote the moments of the transformed variable
$\mathbf{z}' = \mathbf{G}^T\mathbf{z}$ as
$\mub' = \mathbf{G}^T\mub$ and
$\Sigmab_i' = \mathbf{G}^T\Sigmab_i\mathbf{G}$.
Then we have the following relation
\begin{equation}
  \mathbf{H}^T
  \Omegab_i
  \mathbf{H} = 
  \mathbf{H}^T
  \begin{bmatrix}
    \Sigmab_i + \mub \mub^T & \mub \\
    \mub^T & 1
  \end{bmatrix}
  \mathbf{H} = 
  \begin{bmatrix}
    \G^T(\Sigmab_i + \mub \mub^T)\G & \G^T\mub \\
    \G\mub^T & 1
  \end{bmatrix}
  =
  \begin{bmatrix}
    \Sigmab_i' + \mub' \mub'^T & \mub' \\
    \mub'^T & 1
  \end{bmatrix}
  =
  \Omegab_i'
\end{equation}
Following the same reasoning as above, if the filters
are transformed as $\F' = \F\mathbf{G}$, this amounts to
transforming the Calvo-Oller embedding of each class as
$\Omegab_i' = \mathbf{H}^T \Omegab_i \mathbf{H}$,
meaning that the pairwise distances between classes
remain the same.

In sum, in the absence of regularizing noise,
there is an equivalent set of solutions,
given by the sets of filters that span the same subspace.
This can be a useful property, for example because there
is no need to worry about scaling in the data space. But the lack
of a unique solution also makes for less interpretable features
(the interpretable object is the subspace spanned by the filters).

%%%%%%%%%%%%%%%%%%%%%%%%%%%%%%%%%%%%%%%%%%%%%%%%%%%%%%%%%%%%%%%%%%%%%%%%%%%

\subsection{Regularization breaks invariance}

The equivalence of solutions above, however, is eliminated when we introduce
regularization as an additive term $\mathbf{I}\sigma^2$ to the
covariance matrices. We show this for the simpler case of smSQFA,
but the same reasoning applies to SQFA\@.

As described in the Methods section, regularization is introduced
by adding $\mathbf{I}\sigma^2$ to each second-moment matrix (equal
to the covariance matrix in the zero-mean case of smSQFA) in the
feature space. That is
\begin{equation}
  \Psib_i = \F^T \mathbb{E}\left[\mathbf{x}\mathbf{x}^T\right] \F + \mathbf{I}\sigma^2
\end{equation}
This means that if the filters are transformed
as $\F' = \F\mathbf{G}$, it is no longer true that
the regularized second-moment matrices are related by
$\Psib_i' = \mathbf{G}^T\Psib_i\mathbf{G}$.
Rather, there will be a different matrix $\mathbf{G}_i$ satisfying
$\Psib_i' = \mathbf{G}_i^T\Psib_i\mathbf{G}_i$ for each class $i$.
Therefore, the pairwise distances between classes will change,
and the objective function will not be invariant to
invertible linear transformations of the filters.

One consequence of adding regularization is that the solution
is no longer invariant to the scale of the filters.
For filters with small norms,
matrices $\Psib_i = \F^T \Phib_i \F + \mathbf{I}\sigma^2$
will be dominated by the regularization term, and thus
more similar (i.e.\ closer) to each other. Then, the solution will
tend towards filters with infinite norm that make the contribution of
the regularization term negligible. In our case the filters
(i.e.\ each column of $\F$) are constrained to have unit
norm, so this effect is avoided.

Following the same reasoning, the regularization term will
penalize filters that lead to small second-moment matrices, since
these will be dominated by the regularization term, which is
identical across classes. Thus,
regularization will favor the directions in the data space
that lead to larger second-moment matrices. This might be useful
in that it makes the filters more robust to estimation noise along
dimensions with small variance, but it might also mask important
information in directions with low squared values. Also, this makes
the results dependent on the choice of the regularization
parameter $\sigma^2$. Future work should explore the effect of
regularization on the features learned by SQFA, and examine
how to choose the regularization parameter.

The breaking of invariance by regularization has the
desirable effect of making the specific filters learned more
reproducible and therefore more interpretable.

%%%%%%%%%%%%%%%%%%%%%%%%%%%%%%%%%%%%%%%%%%%%%%%%%%%%%%%%%%%%%%%%%%%%%%%%%%%
%%%%%%%%%%%%%%%%%%%%%%%%%%%%%%%%%%%%%%%%%%%%%%%%%%%%%%%%%%%%%%%%%%%%%%%%%%%
%%%%%%%%%%%%%%%%%%%%%%%%%%%%%%%%%%%%%%%%%%%%%%%%%%%%%%%%%%%%%%%%%%%%%%%%%%%

\section{Details of the speed estimation task}
\label{apd:speed}

\subsection{Dataset synthesis}

The speed estimation dataset consists of
synthetic naturalistic videos of surfaces moving at different
frontoparallel speeds that were synthesized following the procedure
described by \cite{burge_optimal_2015}.

Briefly, each video is initially 60-by-60 pixels
and 15 frames long. In this synthesis procedure, 60 pixels
correspond to 1 degree of visual field (roughly equivalent
to the foveal sampling of photoreceptors) and the videos
have a duration of 250 ms (roughly equivalent to the duration of
a fixation in natural viewing).

A video is synthesized by taking a random
patch from an image of a natural scene and moving it horizontally
at a given speed behind a 60-by-60 pixel aperture, for a duration
of 15 frames. The resulting video is then filtered spatiotemporally
to simulate the response of retinal photoreceptors to the video,
following the procedure
described by \cite{herrera-esposito_optimal_2024}. 
Then, the resulting video is downsampled spatially by a factor
of 2, to 30-by-30 pixel frames, and each frame is averaged vertically,
leading to 30 pixel by 15 frame videos (thus, each video can be
represented as a 450 dimensional vector).
Vertically averaging the movies is equivalent to only considering
filters that are vertically-oriented in the original 2D frame
videos.

Then, to simulate further retinal processing, the video is
converted to contrast, by subtracting and dividing by its
mean intensity across pixels and frames. Denoting the resulting contrast
video by $\mathbf{c}$, we apply divisive normalization 
to the video, dividing it by $\sqrt(\|\mathbf{c}\|^2 + n c_{50}^2)$ 
where $c_{50}^2 = 0.045$ is a constant and $n=450$ is the number
of pixels in the video multiplied by the number of frames.
The resulting video simulates the retinal output in response to a naturalistic
speed, as discussed in \cite{burge_optimal_2015,herrera-esposito_optimal_2024}.

Above, we described the process by which a video is synthesized
for a given speed. In the dataset, 41 retinal speeds (i.e.\ classes)
were used, ranging from -6.0 to 6.0 deg/s with 0.3 deg/s intervals.
For each of the 41 retinal speeds, we synthesized 800 different
naturalistic videos, by using a different randomly sampled patch from
a natural scene for each video. This adds nuisance naturalistic variability
to the dataset. We used 500 videos per speed for training (a total of
20500 videos) and 300 videos per speed for testing (a total of 12300 videos).

Note that because the videos are generated with small patches randomly
sampled from natural images, the expected intensity value at each pixel
and frame is approximately the same (because natural image patches are
approximately stationary), independent of speed.
Then, because the videos were constrained to be contrast
videos--i.e.\ they are formed by subtracting off and dividing by the
intensity mean, consistent with early operations in the human
visual system \citep{burge_optimal_2015}--the mean across patches equals zero,
independent of speed.
This results in a dataset where the classes are all approximately zero-mean.

%%%%%%%%%%%%%%%%%%%%%%%%%%%%%%%%%%%%%%%%%%%%%%%%%%%%%%%%%%%%%%%%%%%%%%%%%%%

\subsection{AMA-Gauss training}

The AMA-Gauss model was implemented following the
description in
\cite{jaini_linking_2017,herrera-esposito_optimal_2024}.
First, a set of linear filters $\mathbf{F}$ is applied to each
pre-processed video, and a sample of independent noise 
is added to each filter output. This results in a noisy response
vector $\mathbf{R} = \F^T \mathbf{c} + \lambda$, where
$\lambda \sim \mathcal{N}(0, \mathbf{I}\sigma^2)$ is the added noise.
Then, a QDA-like decoder is used to classify the videos based
on the noisy response vectors, assuming that the noisy response vectors
are Gaussian distributed conditional on the speed (i.e.\ the class).
The decoder computes the likelihood of the response given each class,
that is, $p(\mathbf{R} | y=i)$, and using the priors (which we set as flat)
and Bayes rule, it computes the posterior
distribution $p(y | \mathbf{R})$ for the different speeds.
For a given video $\mathbf{c}$ corresponding to
true class $y=j$, the AMA-Gauss training loss is the negative
log-posterior at the correct class, that is,
$-\log p(y=j | \mathbf{R})$. The filters $\F$ that
minimize the loss are obtained using gradient descent.
The filters (columns of $\F$) are constrained to have
unit norm. We use the same regularization parameter $\sigma^2$ for
AMA-Gauss and for SQFA. AMA-Gauss filters were learned in a
pairwise fashion.

%%%%%%%%%%%%%%%%%%%%%%%%%%%%%%%%%%%%%%%%%%%%%%%%%%%%%%%%%%%%%%%%%%%%%%%%%%%
%%%%%%%%%%%%%%%%%%%%%%%%%%%%%%%%%%%%%%%%%%%%%%%%%%%%%%%%%%%%%%%%%%%%%%%%%%%
%%%%%%%%%%%%%%%%%%%%%%%%%%%%%%%%%%%%%%%%%%%%%%%%%%%%%%%%%%%%%%%%%%%%%%%%%%%

\section{Details of the neural decoding task}
\label{apd:neural}

The response of a neuron in a given trial is the number of spikes that it produces
in between 50 and 100 ms after stimulus onset (we used a fraction of the stimulus
presentation time to avoid the performance ceiling). The recordings in the
dataset were much longer, of 1.28 s, but we used a reduced window
of 50-100 ms to make the task more challenging and to avoid ceiling effects.

We applied a preprocessing step to the dataset to remove outlier trials
and neurons, because covariance estimation can be heavily affected by outliers
and neural datasets can have a considerable number of outlier trials.
We used a two-step procedure. First, we removed trials where the
population response was an outlier. We standardized the population response
for each class separately, and projected the resulting population
response onto the first 10 principal components. We then obtained a
robust covariance estimator using the Minimum Covariance Determinant (MCD)
estimator. Lastly, we removed trials whose Mahalanobis distance to the class mean
was larger than the median Mahalanobis distance plus 4 standard deviations.
This procedure removed 1.2\% of the trials.

Second, we removed neurons with a large number of outlier trials.
For each neuron, we counted the number of trials with a firing rate
deviating more than 3 standard deviations from the class mean. We
removed neurons with more than 60 outlier trials. This procedure removed
3.3\% of the neurons.

\end{document}

%% file: math_commands.tex
\usepackage{amsmath,amsfonts,bm}

\def\eqref#1{equation~\ref{#1}}
\def\1{\bm{1}}

\DeclareMathAlphabet{\mathsfit}{\encodingdefault}{\sfdefault}{m}{sl}
\SetMathAlphabet{\mathsfit}{bold}{\encodingdefault}{\sfdefault}{bx}{n}

\DeclareMathOperator*{\argmax}{arg\,max}

%% file: algorithm.tex
% SQFA optimization algorithm.
% This file assumes the preamble of tmlr/main.tex.

\begin{algorithm}[t]
\caption{SQFA optimization}
\label{alg:sqfa}
\begin{center}
\begin{minipage}{0.98\linewidth}
\hrule
\vspace{0.5ex}

\textbf{Input:} $\{\gamb_i,\Phib_i\}_{i=1}^c$, $m$, $\sigma^2$.
\quad
\textbf{Output:} $\F \in \mathbb{R}^{n \times m}$.

\begin{enumerate}\itemsep0.25ex
  \item Initialize $\eta \in \mathbb{R}^{n \times m}$,
    $\mathcal{L}_{\mathrm{prev}} \gets -\infty$,
    $\Delta \gets \infty$.
  \item While $\Delta \geq 10^{-6}$:
    \begin{enumerate}
      \item $\F \gets g(\eta)$, where $g$ normalizes the columns of $\eta$.
      \item For $i=1,\ldots,c$:
        \par\smallskip
        \hspace*{1.5em}\begin{tabular}{@{}r@{\;}c@{\;}l@{}}
        $\mub_i$ & $\gets$ & $\F^T \gamb_i,$ \\
        $\Sigmab_i$ & $\gets$ & $\F^T \Phib_i \F + \sigma^2 I_m,$ \\
        $\Omegab_i$ & $\gets$ &
          $\begin{bmatrix}
            \Sigmab_i + \mub_i \mub_i^T & \mub_i \\
            \mub_i^T & 1
          \end{bmatrix}$
        \end{tabular}
        \par\smallskip
      \item $\mathcal{L}(\eta) \gets \sum_{i<j} d_{AI}(\Omegab_i,\Omegab_j)$
      \item $\eta \gets \mathrm{L\mbox{-}BFGS}(\eta, \mathcal{L})$
      \item $\Delta \gets
        |\mathcal{L}(\eta)-\mathcal{L}_{\mathrm{prev}}|$; \quad
        $\mathcal{L}_{\mathrm{prev}} \gets \mathcal{L}(\eta)$.
    \end{enumerate}
  \item Return $\F = g(\eta)$.
\end{enumerate}

\vspace{0.25ex}
\hrule
\end{minipage}
\end{center}
\end{algorithm}